\documentclass{article}

\usepackage{microtype}
\usepackage{graphicx}
\usepackage{booktabs} 
\usepackage{enumitem}
\usepackage{multirow}
\usepackage{diagbox}
\usepackage{subcaption}
\usepackage{float}

\newcommand{\keypoint}[1]{\vspace{0.1cm}\noindent\textbf{#1}}

\usepackage{hyperref}

\usepackage[accepted]{icml2023}
\usepackage{amsmath}
\usepackage{amssymb}
\usepackage{mathtools}
\usepackage{amsthm}

\usepackage[capitalize,noabbrev]{cleveref}

\theoremstyle{plain}
\newtheorem{theorem}{Theorem}[section]

\theoremstyle{definition}
\newtheorem{definition}[theorem]{Definition}

\theoremstyle{remark}

\newcommand{\acc}[2]{{\scriptsize{#1}\tiny{$\pm$#2}}}

\usepackage[textsize=tiny]{todonotes}

\icmltitlerunning{A Closer Look at Few-shot Classification Again}

\begin{document}

\twocolumn[
\icmltitle{A Closer Look at Few-shot Classification Again}



\icmlsetsymbol{equal}{*}

\begin{icmlauthorlist}
\icmlauthor{Xu Luo}{equal,uestc}
\icmlauthor{Hao Wu}{equal,uestc}
\icmlauthor{Ji Zhang}{uestc}
\icmlauthor{Lianli Gao}{uestc}
\icmlauthor{Jing Xu}{HIT}
\icmlauthor{Jingkuan Song}{uestc}
\end{icmlauthorlist}

\icmlaffiliation{uestc}{University of Electronic Science and Technology of China}
\icmlaffiliation{HIT}{Harbin Institute of Technology Shenzhen}

\icmlcorrespondingauthor{Jingkuan Song}{jingkuan.song@gmail.com}

\icmlkeywords{Machine Learning, ICML}

\vskip 0.3in
]



\printAffiliationsAndNotice{\icmlEqualContribution} 

\begin{abstract}
    Few-shot classification consists of a training phase where a model is learned on a relatively large dataset and an adaptation phase where the learned model is adapted to previously-unseen tasks with limited labeled samples. In this paper, we empirically prove that the training algorithm and the adaptation algorithm can be completely disentangled, which allows algorithm analysis and design to be done individually for each phase. Our meta-analysis for each phase reveals several interesting insights that may help better understand key aspects of few-shot classification and connections with other fields such as visual representation learning and transfer learning. We hope the insights and research challenges revealed in this paper can inspire future work in related directions. Code and pre-trained models (in PyTorch) are available at \url{https://github.com/Frankluox/CloserLookAgainFewShot}.

\end{abstract}

\section{Introduction}
During the last decade, deep learning approaches have made remarkable progress in large-scale image classification problems~\cite{AlexNet,ResNet}. Since there are infinitely many categories in the real world that cannot be learned at once, a desire following success in image classification is to equip models with the ability to efficiently learn new visual concepts. This demand gives rise to few-shot classification~\cite{fei2006one,MatchingNet}---the problem of learning a model capable of adapting to new classification tasks given only few labeled samples. 

This problem can be naturally broken into two phases: the training phase for learning an adaptable model and the adaptation phase for adapting the model to new tasks. To make quick adaptation possible, it is natural to think that the design of the training algorithm should prepare for the algorithm used for adaptation. 
For this reason, pioneering works \cite{MatchingNet,MAML,op-as-model} formalize the problem with meta-learning framework, where the training algorithm directly aims at optimizing the adaptation algorithm during training in a learning-to-learn fashion. Attracted by meta-learning's elegant formalization and properties well suited for few-shot learning, many subsequent works designed different meta-learning mechanisms to solve few-shot classification problems.

It is then a surprise to find that a simple transfer learning baseline --- learning a supervised model using the training set and adapting it using a simple adaptation algorithm (e.g., logistic regression) --- performs better than all meta-learning methods \cite{closerlook,rethinking,exploring}. Since simple supervised training is not designed specifically for few-shot classification,  this observation reveals that the training algorithm can be designed without considering the choice of adaptation algorithm while achieving satisfactory performance.
In this work, we take a step further and ask the following question:

\emph{Are training and adaptation algorithms completely uncorrelated in few-shot classification?}

Here, ``completely uncorrelated'' means that  the performance ranking of any set of adaptation algorithms is not affected by the choice of training algorithms, and vice versa. If this is true, the problem of finding the best combinations of training and adaptation algorithms can be reduced to optimizing the training and adaptation algorithms individually, which may largely ease the algorithm design process in the future. We give an affirmative answer to this question by conducting a systematic study on a variety of training and adaptation algorithms used in few-shot classification. 

This ``uncorrelated'' property also offers us an opportunity to independently analyze algorithms of one phase by fixing the algorithm of the other phase. We conduct such analysis in Section \ref{sec-train} for training algorithms and Section \ref{sec-test} for adaptation algorithms. By varying important factors like dataset scale, model architectures for the training phase, and shots, ways, data distribution for the adaptation phase, we obtain several interesting observations that lead to a deeper understanding of few-shot classification and reveal some critical relations to visual representation learning and transfer learning literature. Such meta-level understanding can be useful for future few-shot learning research. 
The analysis for each phase leads to the following key observations:

\begin{enumerate}[leftmargin=*] 
    \item We observed a different neural scaling law in few-shot classification that test error falls off as a power law with the number of training classes, instead of the number of training samples per class. This observation highlights the importance of the number of training classes in few-shot classification and may help future research further understand the crucial difference between few-shot classification and other vision tasks.
    

    %

\item 
We found two evaluated datasets on which increasing the scale of training dataset does not always lead to better few-shot performance. This suggests that it is never realistic to train a model that can solve all possible tasks well just by feeding it a very large amount of data. This also indicates the importance of properly filtering training knowledge for different few-shot classification tasks.



\item We found that standard ImageNet performance is not a good predictor of few-shot performance for supervised models (contrary to previous observations in other vision tasks), but it does predict well for self-supervised models. This observation may become  the  key to understanding both the difference between few-shot classification and other vision tasks, and the difference between supervised learning and self-supervised learning.



\item We found that, contrary to a common belief that fine-tuning the whole network with few samples would lead to severe overfitting,   vanilla fine-tuning performs the best among all adaptation algorithms even when data is extremely scarce, e.g., 5-way 1-shot task. In particular, partial finetune methods that are designed to overcome the overfitting problem of vanilla finetune in few-shot setting perform worse. The advantage of finetune expands with the increase of the number of ways, shots and the degree of task distribution shift. However, finetune methods suffer from extremely high time complexity. We show that the difference in these factors is the reason why state-of-the-art methods in different few-shot classification benchmarks differ in adaptation algorithms.



\end{enumerate}





\section{The Problem of Few-shot Classification}
 Few-shot classification aims to learn a model that can quickly adapt to a novel classification task given only few observations. In the training phase, given a training dataset $\mathcal{D}^{train}=\{x_n,y_n\}_{n=1}^{|\mathcal{D}^{train}|}$ with $N_C$ classes, where $x_i\in\mathbb{R}^D$ is the $i$-th image and $y_i\in[N_C]$ is its label, a model $f_{\theta}$ is learned via a training algorithm $\mathcal{A}^{train}$, i.e., $\mathcal{A}^{train}(\mathcal{D}^{train})=f_{\theta}$. In the adaptation phase, a series of  few-shot classification tasks $\mathcal{T}=\{\tau_i\}_{i=1}^{N_\mathcal{T}}$ are constructed from the test dataset $\mathcal{D}^{test}$ with classes and domains possibly different from those of $\mathcal{D}^{train}$. Each task $\tau$ consists of a support set $\mathcal{S}=\{(x_i,y_i)\}_{i=1}^{N_\mathcal{S}}$ used for adaptation and a query set $\mathcal{Q}=\{(x_i^*,y_i^*)\}_{i=1}^{N_\mathcal{Q}}$ that is used for evaluation and shares the same label space with $\mathcal{S}$. $\tau$ is called a $N$-way $K$-shot task if there are $N$ classes in the support set $\mathcal{S}$ and each class contains exactly $K$ samples. To solve each task $\tau$, the adaptation algorithm $\mathcal{A}^{adapt}$ takes the learned model $f_{\theta}$ and the support set $\mathcal{S}$ as inputs, and produces a new classifier $g(\cdot;f_\theta,\mathcal{S}):\mathbb{R}^D\rightarrow[N]$. The constructed classifier will be evaluated on the query set $\mathcal{Q}$ to test its generalization ability. The evaluation metric is the average performance over all sampled tasks. We denote both the resultant average accuracy and the radius of 95\% confidence interval as a function of training and adaptation algorithms:  $\mathrm{Avg}(\mathcal{A}^{train},\mathcal{A}^{adapt})$ and $\mathrm{CI}(\mathcal{A}^{train},\mathcal{A}^{adapt})$, respectively.

 Depending on the form of training algorithm $\mathcal{A}^{train}$, the model $f_\theta$ can be different. For non-meta-learning methods, $f_\theta:\mathbb{R}^D\rightarrow\mathbb{R}^d$ is simply a feature extractor that takes an image $x\in\mathbb{R}^D$ as input and outputs a feature vector $z\in\mathbb{R}^d$. Thus any visual representation learning algorithms can be used as $\mathcal{A}^{train}$. For meta-learning methods, the training algorithm directly aims at optimizing the performance of the adaptation algorithm $\mathcal{A}^{adapt}$ in a learning-to-learn fashion. Specifically, meta-learning methods firstly parameterize the adaptation algorithm $\mathcal{A}^{adapt}_\theta$ that makes it optimizable. Then the model $f_\theta$ used for training is set equal to $\mathcal{A}^{adapt}_\theta$, i.e., $\mathcal{A}^{train}(\mathcal{D}^{train})=f_\theta=\mathcal{A}^{adapt}_\theta$. The training process consists of constructing pseudo few-shot classification tasks $\mathcal{T}^{train}=\{(\mathcal{S}_t^{train},\mathcal{Q}_t^{train})\}_{t=1}^{N_\mathcal{T}^{train}}$ from $\mathcal{D}^{train}$ that take the same form with tasks during adaptation. In each iteration $t$, just like what will be done in the adaptation phase, the model $f_\theta$ takes $\mathcal{S}_t^{train}$ as input and outputs a classifier $g(\cdot;\mathcal{S}_t)$. Images in $\mathcal{Q}_t^{train}$ are then fed into  $g(\cdot;\mathcal{S}_t)$ and return a loss that is used to update $f_\theta$. After training, $f_\theta$ is directly used as the adaptation algorithm $\mathcal{A}^{adapt}_\theta$. Although different from non-meta-learning methods, most meta-learning algorithms still set the learnable parameters $\theta$ as the parameters of a feature extractor, making it possible to change the algorithm used for adaptation.
 


 
\begin{table*}[t]
\scriptsize
\centering
\setlength\tabcolsep{5pt}
\caption{Few-shot classification performance of pairwise combinations of a variety of training and adaptation algorithms. All evaluation tasks are 5-way 5-shot tasks sampled from Meta-Dataset (excluding ImageNet). We sample 2000 tasks per dataset in Meta-Dataset and report the average accuracy over all datasets along with the 95\% confidence interval. The algorithms are listed according to their partial order according to Definition~\ref{uncorrelatedness} from top to bottom and from left to right.  * denotes training algorithm that uses transductive BN~\cite{metaBN} that produces a much higher, unfair performance using Fintune  and TSA as adaptation algorithms. $\dag$: TSA and eTT are both architecture-specific partial-finetune algorithms, thus TSA can be used for CNN only and eTT for original ViT only.}
\label{correlation}
\renewcommand{\arraystretch}{1.2}
    \begin{tabular}{ccccccccccc}
    \hline
    & & & 
    \multicolumn{8}{c}{Adaptation algorithm} \\ 
    
    Training algorithm & Training dataset & Architecture
                                & MatchingNet                            & MetaOpt                       & NCC                           & LR                            & URL                           & CC                            & TSA/eTT$^\dag$                           & Finetune                 \\
    \hline

    PN  & \emph{mini}ImageNet & Conv-4                        & \acc{48.54}{0.4}          & \acc{49.84}{0.4}          & \acc{51.38}{0.4}          & \acc{51.65}{0.4}          & \acc{51.82}{0.4}          & \acc{51.56}{0.4}          & \acc{58.08}{0.4}          & \acc{60.88}{0.4}            \\
    MAML$^*$ & \emph{mini}ImageNet   & Conv-4                    & \acc{53.71}{0.4}          & \acc{53.69}{0.4}          & \acc{55.01}{0.4}          & \acc{55.03}{0.4}          & \acc{55.66}{0.4}          & \acc{55.63}{0.4}          & \acc{62.80}{0.4}          & \acc{64.87}{0.4}   \\
    CE   & \emph{mini}ImageNet   & Conv-4                    & \acc{54.68}{0.4}          & \acc{56.79}{0.4}          & \acc{58.54}{0.4}          & \acc{58.26}{0.4}          & \acc{59.63}{0.4}          & \acc{59.20}{0.5}          & \acc{64.14}{0.4}          & \acc{65.12}{0.4}   \\
    MatchingNet   & \emph{mini}ImageNet   & ResNet-12                   & \acc{55.62}{0.4}          & \acc{57.20}{0.4}          & \acc{58.91}{0.4}          & \acc{58.99}{0.4}          & \acc{61.20}{0.4}          & \acc{60.50}{0.4}          & \acc{64.88}{0.4}          & \acc{67.93}{0.4} \\
    MAML$^*$  & \emph{mini}ImageNet    & ResNet-12                  & \acc{58.42}{0.4}          & \acc{58.52}{0.4}          & \acc{59.65}{0.4}          & \acc{60.04}{0.4}          & \acc{60.38}{0.4}          & \acc{60.50}{0.4}          & \acc{71.15}{0.4}          & \acc{73.13}{0.4}   \\
    PN       & \emph{mini}ImageNet   & ResNet-12                & \acc{60.19}{0.4}          & \acc{61.70}{0.4}          & \acc{63.71}{0.4}          & \acc{64.46}{0.4}          & \acc{65.64}{0.4}          & \acc{65.76}{0.4}          & \acc{70.44}{0.4}          & \acc{74.23}{0.4}  \\
    MetaOpt  & \emph{mini}ImageNet  & ResNet-12                 & \acc{62.06}{0.4}          & \acc{63.94}{0.4}          & \acc{65.81}{0.4}          & \acc{66.03}{0.4}          & \acc{67.47}{0.4}          & \acc{67.24}{0.4}          & \acc{72.07}{0.4}          & \acc{74.96}{0.4}   \\
    DeepEMD & \emph{mini}ImageNet  & ResNet-12                 & \acc{62.67}{0.4} & \acc{64.15}{0.4} & \acc{66.14}{0.4} & \acc{66.14}{0.4} & \acc{68.66}{0.4} & \acc{69.76}{0.4} & \acc{74.21}{0.4} & \acc{74.83}{0.4}\\
    CE      & \emph{mini}ImageNet   & ResNet-12                 & \acc{63.27}{0.4}          & \acc{64.91}{0.4}          & \acc{66.96}{0.4}          & \acc{67.14}{0.4}          & \acc{69.78}{0.4}          & \acc{69.52}{0.4}          & \acc{74.30}{0.4}          & \acc{74.89}{0.4}    \\
    Meta-Baseline     & \emph{mini}ImageNet   & ResNet-12                  & \acc{63.25}{0.4}          & \acc{65.02}{0.4}          & \acc{67.28}{0.4}          & \acc{67.56}{0.4}          & \acc{69.84}{0.4}          & \acc{69.76}{0.4}          & \acc{73.94}{0.4}          & \acc{75.04}{0.4}   \\
    COS      & \emph{mini}ImageNet  & ResNet-12                 & \acc{63.99}{0.4}          & \acc{66.09}{0.4}          & \acc{68.31}{0.4}          & \acc{69.26}{0.4}          & \acc{70.71}{0.4}          & \acc{71.03}{0.4}          & \acc{75.10}{0.4}          & \acc{75.68}{0.4}  \\

        PN     & ImageNet  & ResNet-50                   & \acc{63.68}{0.4}          & \acc{65.79}{0.4}          & \acc{68.40}{0.4}          & \acc{68.87}{0.4}          & \acc{69.69}{0.4}          & \acc{70.81}{0.4}          & \acc{74.15}{0.4}          & \acc{78.42}{0.4}         \\
    S2M2 & \emph{mini}ImageNet & WRN-28-10 & \acc{64.41}{0.4} & \acc{66.59}{0.4} & \acc{68.67}{0.4} & \acc{69.16}{0.4} & \acc{70.88}{0.4} & \acc{71.38}{0.4} & \acc{74.94}{0.4} & \acc{76.89}{0.4} \\ 
    FEAT & \emph{mini}ImageNet & 
    ResNet-12 & \acc{65.42}{0.4} & \acc{67.15}{0.4} & \acc{69.06}{0.4} & \acc{69.21}{0.4} & \acc{71.24}{0.4} & \acc{72.07}{0.4} & \acc{75.99}{0.4} & \acc{76.83}{0.4} \\
    IER    & \emph{mini}ImageNet   & ResNet-12                  & \acc{65.37}{0.4}          & \acc{67.31}{0.4}          & \acc{69.30}{0.4}          & \acc{70.01}{0.4}          & \acc{72.48}{0.4}          & \acc{72.85}{0.4}          & \acc{76.70}{0.4}          & \acc{77.54}{0.4}       \\

    Moco v2    & ImageNet   & ResNet-50              & \acc{65.47}{0.5}          & \acc{68.63}{0.4}          & \acc{71.05}{0.4}          & \acc{71.49}{0.4}          & \acc{74.46}{0.4}          & \acc{74.57}{0.4}          & \acc{79.70}{0.4}          & \acc{79.98}{0.4}      \\
    Exemplar v2    & ImageNet   & ResNet-50              & \acc{67.70}{0.5}          & \acc{70.07}{0.4}          & \acc{72.55}{0.4}          & \acc{72.93}{0.4}          & \acc{75.26}{0.4}          & \acc{76.83}{0.4}          & \acc{80.22}{0.4}          & \acc{81.75}{0.4}      \\
    DINO     & ImageNet    & ResNet-50               & \acc{73.97}{0.4}          & \acc{76.45}{0.4}          & \acc{78.30}{0.4}          & \acc{78.72}{0.4}          & \acc{80.73}{0.4}          & \acc{81.05}{0.4}          & \acc{83.64}{0.4}          & \acc{83.20}{0.4}   \\
    CE      & ImageNet   & ResNet-50                 & \acc{74.75}{0.4}          & \acc{76.94}{0.4}          & \acc{78.96}{0.4}          & \acc{79.57}{0.4}          & \acc{80.89}{0.4}          & \acc{81.51}{0.4}          & \acc{84.07}{0.4}          & \acc{84.92}{0.4}  \\
    BiT-S    & ImageNet   & ResNet-50                & \acc{75.44}{0.4}          & \acc{77.86}{0.4}          & \acc{79.84}{0.4}          & \acc{79.97}{0.4}          & \acc{81.79}{0.4}          & \acc{81.91}{0.4}          & \acc{84.84}{0.3}          & \acc{86.40}{0.3}   \\

    CE  & ImageNet & Swin-B & \acc{75.17}{0.4}          & \acc{77.81}{0.4}          & \acc{80.06}{0.4}          & \acc{81.04}{0.4}          & \acc{82.55}{0.4}          & \acc{82.46}{0.4}          & -         & \acc{88.16}{0.3}   \\
    DeiT    & ImageNet & ViT-B   & \acc{75.82}{0.4}          & \acc{78.34}{0.4}          & \acc{80.62}{0.4}          & \acc{81.68}{0.4}          & \acc{82.80}{0.3}          & \acc{83.13}{0.4}          & \acc{84.22}{0.3}                      & \acc{87.62}{0.3}       \\
    CE  & ImageNet  & ViT-B  & \acc{76.78}{0.4}          & \acc{78.81}{0.4}          & \acc{80.65}{0.4}          & \acc{81.13}{0.3}          & \acc{82.69}{0.3}          & \acc{82.77}{0.3}          & \acc{85.60}{0.3}                   & \acc{88.48}{0.3}         \\
    DINO    & ImageNet  & ViT-B                  & \acc{76.44}{0.4}          & \acc{79.11}{0.4}          & \acc{81.23}{0.4}          & \acc{82.01}{0.4}          & \acc{84.16}{0.3}          & \acc{84.44}{0.3}          & \acc{86.25}{0.3}                         & \acc{88.04}{0.3}   \\
    CLIP    & WebImageText  & ViT-B             & \acc{78.06}{0.4}          & \acc{81.20}{0.4}          & \acc{83.04}{0.3}          & \acc{83.22}{0.3}          & \acc{84.11}{0.3}          & \acc{84.20}{0.3}          & \acc{87.66}{0.3}                         & \acc{90.26}{0.3}   \\
    \hline
    \end{tabular}
\end{table*}


\section{Are Training and Adaptation Algorithms Uncorrelated?}
\label{sec-correlation}
Given a set of training algorithms $M^{train}=\{\mathcal{A}^{train}_i\}_{i=1}^{m_1}$ and a set of adaptation algorithms $M^{adapt}=\{\mathcal{A}^{adapt}_i\}_{i=1}^{m_2}$, we say that $M^{train}$ and $M^{adapt}$ are uncorrelated if changing algorithms from  $M^{train}$ does not influence the performance ranking of algorithms from $M^{adapt}$, and vice versa. To give a precise description, we first define a partial order.
\begin{definition}
\label{relative relation}
We say two training algorithms $\mathcal{A}^{train}_a,\mathcal{A}^{train}_b$ have the partial order $\mathcal{A}^{train}_a\preceq\mathcal{A}^{train}_b$, if for all $i\in[m_2]$,
\begin{align}
&\mathrm{Avg}(\mathcal{A}^{train}_a,\mathcal{A}^{adapt}_i)-\mathrm{CI}(\mathcal{A}^{train}_a,\mathcal{A}^{adapt}_i)\nonumber\\<&\mathrm{Avg}(\mathcal{A}^{train}_b,\mathcal{A}^{adapt}_i)+\mathrm{CI}(\mathcal{A}^{train}_b,\mathcal{A}^{adapt}_i).
\end{align}
\end{definition}
 This inequality holds when the values inside the confidence interval of $\mathcal{A}^{train}_b$ are all larger than or at least have overlap with that of $\mathcal{A}^{train}_a$ when evaluated with every adaptation algorithm in $M^{adapt}$. This implies that there is a considerable probability that the performance of $\mathcal{A}^{train}_b$ is no worse than $\mathcal{A}^{train}_a$ when combined with any possible adaptation algorithm $\mathcal{A}^{adapt}_i$, thus the ranking of the two training algorithms are not influenced by adaptation algorithms with high probability. We here use $\preceq$ instead of $\prec$ to show that the defined partial order is not strict, so it is valid that $\mathcal{A}^{train}_a \preceq \mathcal{A}^{train}_b$ and $\mathcal{A}^{train}_b \preceq \mathcal{A}^{train}_a$ hold simultaneously, meaning that the two algorithms are comparable. The partial order inside $M^{adapt}$ can be similarly defined by 
exchanging training and adaptation algorithms above. We are now ready to define what it means for two sets of algorithms to be uncorrelated.
\begin{definition}
    \label{uncorrelatedness}
    $M^{train}$ and $M^{adapt}$ are uncorrelated if they are both ordered sets wrt the partial order relation defined in Definition \ref{relative relation}.\footnote{The partial order in Definition \ref{relative relation} may not satisfy transitivity, i.e., if $\mathcal{A}^{train}_1\preceq\mathcal{A}^{train}_2$ and $\mathcal{A}^{train}_2\preceq\mathcal{A}^{train}_3$, it is possible that  $\mathcal{A}^{train}_1\preceq\mathcal{A}^{train}_3$ does not hold. However, in our experiment, such cases do not exist. Thus we assume the transitivity holds and we can always get an ordered set of algorithms from one-by-one relations.}
\end{definition}

Now, to see whether training and adaptation algorithms in few-shot classification are uncorrelated, we choose a wide range of training and adaptation algorithms from previous few-shot classification methods with various training datasets and network architectures to form $M^{train}$ and $M^{adapt}$. We then conduct experiments on each pair of algorithms, one from 
$M^{train}$ and another from $M^{adapt}$, to check whether the two sets are ordered sets.

\keypoint{Algorithms evaluated.} The selected set of training algorithms $M^{train}$ encompasses both meta-learning and non-meta-learning methods. For meta-learning methods, we evaluate MAML~\cite{MAML}, ProtoNet~\cite{PN}, MatchingNet~\cite{MatchingNet}, MetaOpt~\cite{metaopt}, Feat~\cite{ye2020few}, DeepEMD~\cite{deepemd} and MetaBaseline~\cite{metabaseline}. For non-meta-learning methods, we evaluate supervised algorithms including Cross-Entropy baseline~\cite{closerlook}, COS~\cite{COSOC}, S2M2~\cite{mangla2020charting}, IER~\cite{exploring}, BiT~\cite{bit}, Exemplar v2~\cite{Exemplar} and DeiT~\cite{DeiT}; unsupervised algorithms including MoCo-v2~\cite{MoCo} and DINO~\cite{DINO}; and a multimodal pre-training algorithm CLIP~\cite{CLIP}. $M^{adapt}$ encompasses the ones from meta-learning methods including MatchingNet, MetaOpt, Nearest Centroid Classifier (PN), Finetune (MAML); the ones from non-meta-learning methods including Logistic Regression~\cite{rethinking}, URL~\cite{URL}, Cosine Classifier~\cite{closerlook}; and test-time-only methods TSA~\cite{TSA} and eTT~\cite{eTT}. 

\keypoint{Datasets.} For the test dataset, we choose Meta-Dataset~\cite{metadataset}, a dataset of datasets that covers 10 diverse vision datasets from different domains. We remove ImageNet from Meta-Dataset to avoid label leakage from training. For training, we choose three datasets of different scales: the train split of \emph{mini}ImageNet~\cite{MatchingNet} that contains 38400 images from 64 classes,
the train split of ImageNet~\cite{imagenet} that contains more than 1 million images from 1000 classes, and a large-scale multimodal dataset WebImageText~\cite{CLIP} that contains 400 million (image, text)
pairs. For completeness, we also show traditional \emph{mini}ImageNet-only experiments in Table \ref{correlation-3}-\ref{correlation-4} in the Appendix.

\keypoint{Results.} Table \ref{correlation} shows 5-way 5-shot performance of all pairwise combinations of algorithms from $M^{train}$ and $M^{adapt}$. As seen,  both training algorithms and adaptation algorithms form ordered sets according to Definition \ref{uncorrelatedness}: when we fix any adaptation algorithm (a column in the table), the performance is monotonically increasing (or at least confidence intervals are intersected) as we move from top to bottom; similarly, adaptation algorithms form an ordered set from left to right. 1-shot results are similar and are given in Table \ref{correlation-2} in the Appendix. Since we have covered a bunch of representative few-shot classification algorithms, we can say that with high probability, training and adaptation algorithms are uncorrelated in few-shot classification.

\keypoint{Remark.} According to Definition \ref{uncorrelatedness}, since $M^{train}$ and $M^{adapt}$ are uncorrelated, changing algorithms either in $M^{train}$ or $M^{adapt}$ along the sequences in the ordered set always leads to performance improvement. Thus a simple greedy search on either side of algorithms always leads to global optima. A direct consequence is that, if two phases of algorithms are optimal on their own, their combinations are optimal too. For example, from Table \ref{correlation} we can see that, for 5-way 5-shot tasks on Meta-Dataset, CLIP and Finetune are the optimal training and adaptation algorithms respectively, and their combination also becomes the optimal combination. 

This algorithm-disentangled property would greatly simplify the algorithm design process in few-shot classification. In the next two sections, we will, for the first time, individually analyze each of the two phases of algorithms while fixing the algorithms in the other phase.

\section{Training Analysis}
\label{sec-train}
Throughout this section, we will fix the adaptation algorithm to the Nearest-Centroid Classifier, and analyze some aspects of interest in the training process of few-shot classification. According to Section \ref{sec-correlation}, observations would not change with high probability if we change adaptation algorithms.

\subsection{On the Scale of Training Dataset}
\label{sec-train-scale}
We are first interested in understanding how the scale of training dataset influences few-shot classification performance. In few-shot classification, since classes in training and adaptation do not need to overlap, in addition to increasing the number of samples per class, we can also increase the training dataset size by increasing the number of training classes. This is different from standard vision classification tasks where studying the effect of increasing the number of samples per class is of more interest.

\begin{figure*}[t]
\centering
\centerline{\includegraphics[width=1.0\linewidth]{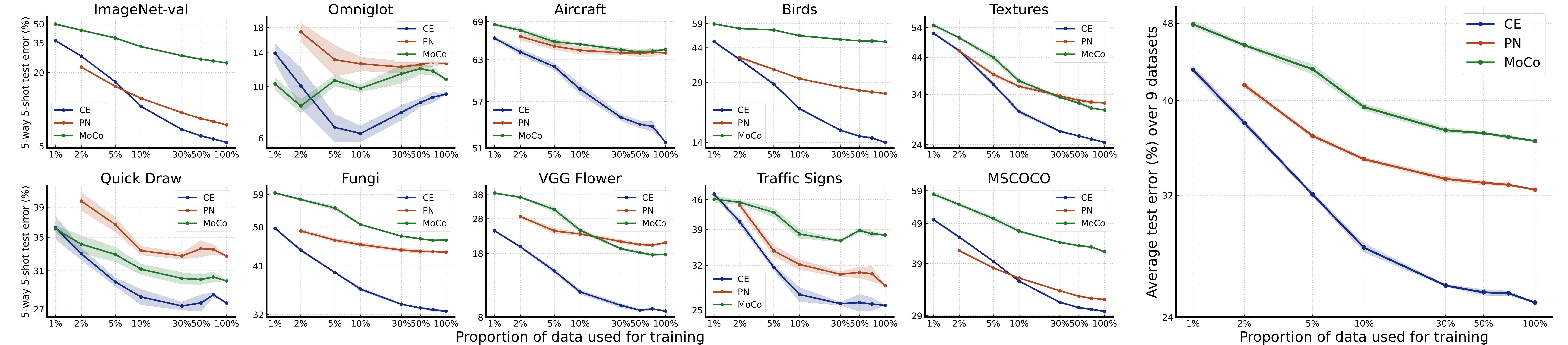}}
\caption{The effect of sample size per training class on few-shot classification performance. We use all $1000$ classes of the training set of ImageNet for training. Both axes are logit-scaled. ImageNet-val means conducting few-shot classification on the original validation set of ImageNet. The average performance is obtained by averaging performance on 9 datasets excluding ImageNet-val. Best viewed in color.}

\label{train-dataset-analysis}

\end{figure*}

\begin{figure*}[t]
\centering
\centerline{\includegraphics[width=1.0\linewidth]{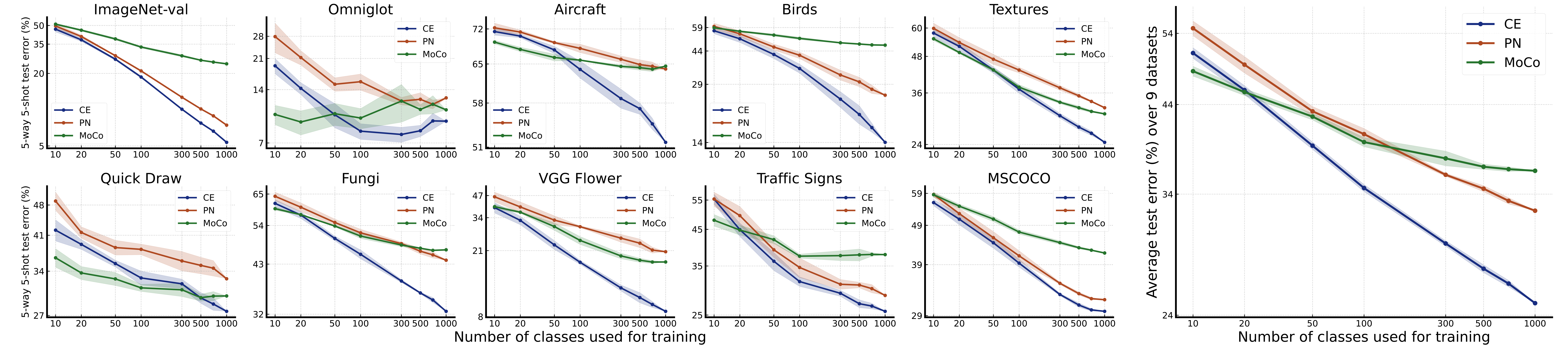}}
\caption{The effect of the number of training classes on few-shot classification performance. For each randomly selected class in ImageNet, we use all of its samples from the training set for training. Both axes are logit-scaled.  Best viewed in color.}

\label{train-dataset-analysis-label}

\end{figure*}

We conduct both types of scaling experiments on the training set of ImageNet, a standard vision dataset that is always used as a pre-training dataset for downstream tasks. We choose three representative training algorithms that cover main types of algorithms: (1) Cross Entropy (CE) training, the standard supervised training in image classification tasks; (2) ProtoNet (PN), a widely-used meta-learning algorithm; (3) MoCo-v2, a strong unsupervised visual representation learning algorithm. For each dataset scale, we randomly select samples or classes 5 times, train a model using the specified training algorithm, and report the average performance and the standard variation over the 5 trials of training. The adaptation datasets we choose include 9 datasets from Meta-Dataset and the standard validation set of ImageNet.  We plot the results of ranging the number of samples per class in Figure \ref{train-dataset-analysis} and the results of ranging the number of classes in Figure \ref{train-dataset-analysis-label}. Both axes are plotted in log scale. We also report the results evaluated on additional 9 datasets in BSCD-FSL benchmark and DomainNet in Figure \ref{additional-train-dataset-analysis}-\ref{additional-train-dataset-analysis-label} in the Appendix. We make the following observations.

\keypoint{Neural scaling laws for training.} Comparing Figure \ref{train-dataset-analysis} and \ref{train-dataset-analysis-label}, we can see that for supervised models (CE and PN), increasing the number of classes is much more effective than increasing the number of samples per class (We give clearer comparisons in Figure \ref{CE_compare}-\ref{MoCo_compare} in the Appendix). The effect of increasing the number of samples per class plateaus quickly, while increasing the number of classes leads to very stable performance improvement throughout all scales. We notice that most performance curves of PN and CE in Figure \ref{train-dataset-analysis-label} look like a straight line. In Figure \ref{LinearFit_CE}-\ref{LinearFit_MoCo} in the appendix we plot the linear fit which verifies our observations. In fact, the Pearson coefficient between the log scale of the number of training classes and the log scale of average test error is $-0.999$ for CE and $-0.995$ for PN, showing strong evidence of linearity. This linearity indicates the existence of a form of neural scaling laws in few-shot classification: test error falls off as a power law with the number of training classes, which is different from neural scaling laws observed in other machine learning tasks \cite{scaling-law-1,scaling-law-2,scaling-law-3} that test error falls off as a power law with the number of training samples per class. Such a difference reveals the intrinsic difference between few-shot classification and other tasks: while seeing more samples in a training class does help in identifying new samples in the same class, it may not help that much in identifying previously-unseen classes in a new task. On the other hand, seeing more classes may help the model learn more potentially useful knowledge that might help distinguish new classes. 

\begin{figure*}[t]
\centering
\centerline{\includegraphics[width=1.0\linewidth]{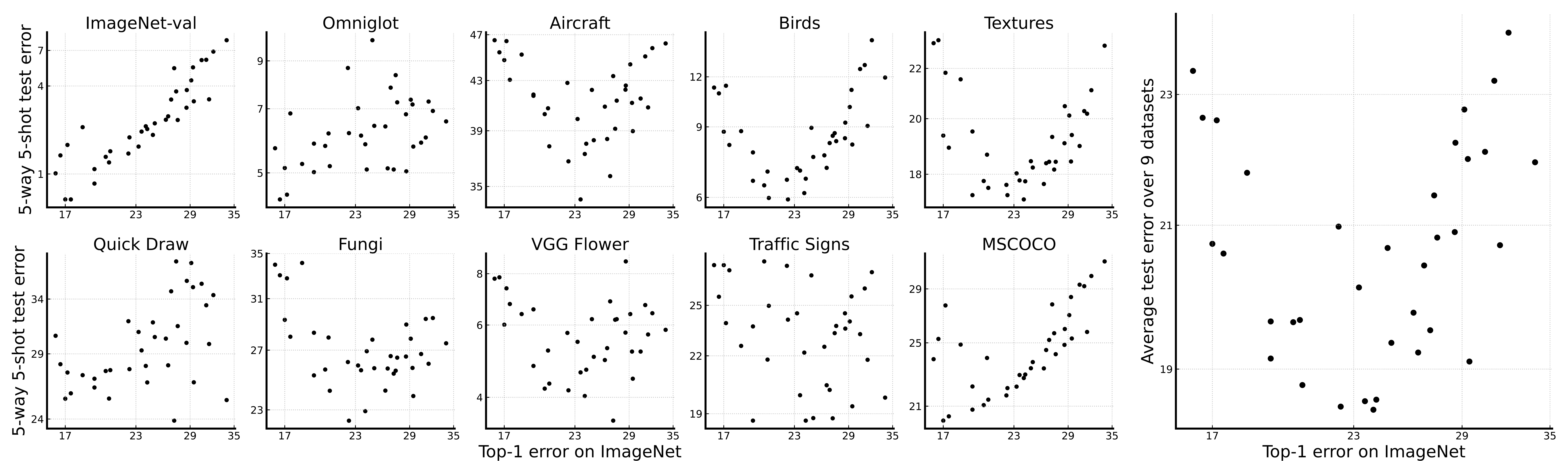}}
\caption{For supervised models, ImageNet performance is not a good predictor of  few-shot classification performance. Each point in a plot refers to a supervised CE model with a specific backbone architecture. Both axes are logit-scaled.}

\label{trainnormal_testfsl}

\end{figure*}

\begin{figure*}[t]
\centering
\centerline{\includegraphics[width=1.0\linewidth]{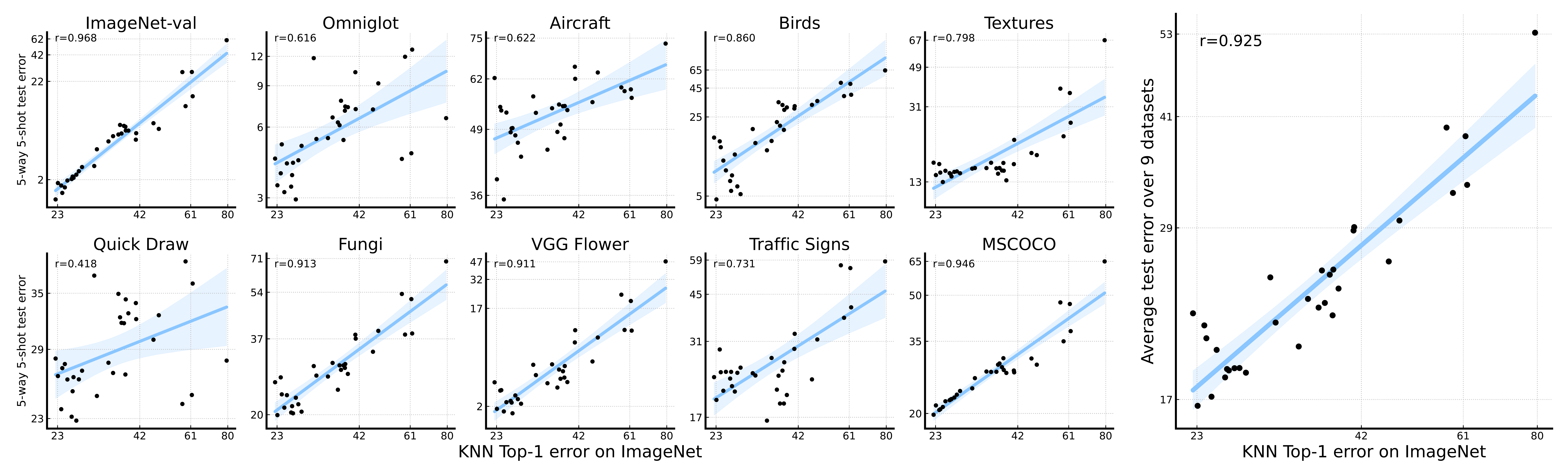}}
\caption{For self-supervised models, ImageNet performance is a good predictor of  few-shot classification performance. Each point in a plot refers to a self-supervised model with a specific training algorithm/architecture. Both axes are logit-scaled. The regression line and a 95\%
confidence interval are plotted in blue. ``r'' refers to the correlation coefficient between the two axes of data.}

\label{trainSSL_testfsl}

\end{figure*}

\keypoint{Bigger is not necessarily better.}
On most evaluated datasets, test error decreases with more training samples/classes. However, on Omniglot and ISIC (shown in Figure \ref{additional-train-dataset-analysis}-\ref{additional-train-dataset-analysis-label}), the error first goes down and then goes up, especially for supervised models. On the contrary, previous works \cite{PN} have shown that a simple PN model, both training and evaluating on Omniglot (class separated), can easily obtain a near-zero error. This indicates that as the number of training samples/classes increases, there exists a progressively larger mismatch between the knowledge learned from ImageNet and the knowledge needed for distinguishing new classes in these two datasets. Thus training a large model on a big dataset that can solve every possible task well is not a realistic hope, unless the training dataset already contains all possible tasks. How to choose a part of the training dataset to train a model on, or how to select positive/useful knowledge from a learned model depending on only a small amount of data in the specified adaptation scenario is an important research direction in few-shot classification.

\keypoint{CE training scales better.} As seen from both figures, PN and MoCo perform comparably to CE on small-scale training data, but as more training data comes in, the gap gradually widens. Considering that all algorithms have been fed with the same amount of data during training, we can infer that CE training indeed scales better than PN and MoCo. This trend seems to be more obvious for fine-grained datasets including Aircraft, Birds, Fungi, and VGG Flower. While this phenomenon needs further investigation, we speculate it is due to that CE simultaneously differentiates all classes during training which requires distinguishing all possible fine-grained classes. On the contrary, meta-learning algorithms like PN typically need to  distinguish only a limited number of classes during each iteration, and self-supervised models like MoCo do not use labels, thus focusing more on global information in images \cite{Exemplar} and performing not well on fine-grained datasets. We leave it for future work to verify if this conjecture holds generally.

\begin{figure*}[t]
\centering
\centerline{\includegraphics[width=1.0\linewidth]{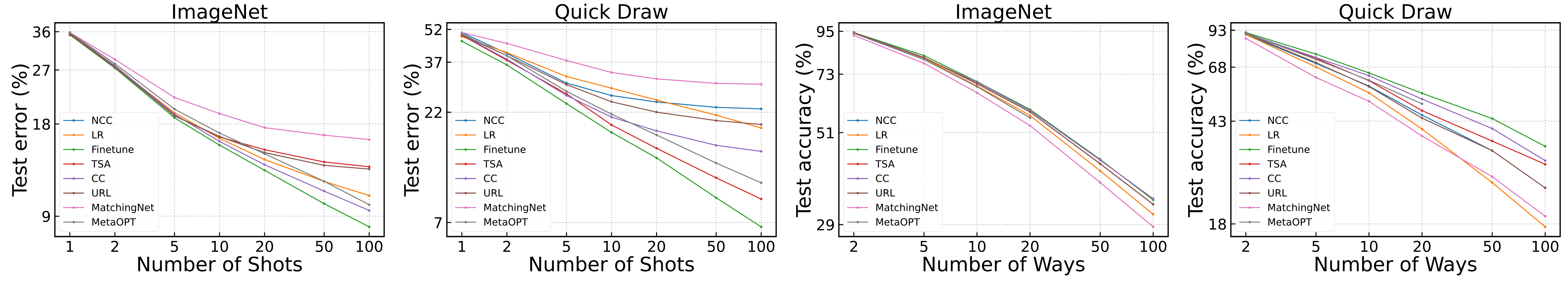}}
\caption{Way and shot experiments of adaptation algorithms on ImageNet and Quick Draw. For shot experiment, we fix the number of ways to 5 and show test error, and for way experiment, we fix the number of shots to 5 and show test accuracy. Both axes are logit-scaled. Best viewed in color.}

\label{test_comparison}

\vskip -0.15in
\end{figure*}

\subsection{ImageNet Performance vs Few-shot Performance}
\label{vs-exp}
Then we fix the scale of training dataset and investigate how the changes in training algorithms and network architectures influence few-shot performance. We especially pay our attention to CE-trained and self-supervised models due to their superior performance shown in Table \ref{correlation}.
Previous studies have revealed that the standard ImageNet performance of CE models trained on ImageNet is a strong predictor (with a linear relationship) of its performance on a range of vision tasks, including transfer learning \cite{transfer_better}, open-set recognition \cite{closed-set} and domain generalization \cite{natural-distribution-shifts}. We here ask if this observation also holds for few-shot classification. If this is true, we can improve few-shot classification on benchmarks that use ImageNet as the training dataset like Meta-Dataset, by just waiting for state-of-the-art ImageNet models. For this, we test 36 pre-trained supervised CE models with different network architectures, including VGG~\cite{VGG}, ResNet~\cite{ResNet}, MobileNet~\cite{MobileNet}, RegNet~\cite{RegNet}, DenseNet~\cite{DenseNet}, ViT~\cite{ViT}, Swin Transformer~\cite{swin} and ConvNext~\cite{ConvNext}. We also test 32 self-supervised ImageNet pre-trained models with different network algorithms and architectures. The algorithms include MoCo-v2~\cite{MoCo}, MoCo-v3~\cite{MoCo-v3}, InstDisc~\cite{InscDisc}, BYOL~\cite{BYOL}, SwAV~\cite{swav}, OBoW~\cite{obow}, SimSiam~\cite{SimSiam}, Barlow Twins~\cite{Barlow}, DINO~\cite{DINO}, MAE~\cite{MAE}, iBOT~\cite{ibot} and EsViT~\cite{EsViT}. We use KNN \cite{DINO} to compute top-1 accuracy for these self-supervised models. We plot results for supervised models in Figure \ref{trainnormal_testfsl} and self-supervised models in Figure \ref{trainSSL_testfsl}.

\keypoint{Supervised ImageNet models overfit to ImageNet performance.}  For supervised models, we can observe from Figure \ref{trainnormal_testfsl} that on most datasets sufficiently different from ImageNet, such as Aircraft, Birds, Textures, Fungi, and VGG Flower, the test error of few-shot classification first decreases and then increases with the improvement of ImageNet performance. The critical point is at about 23\% Top-1 error on ImageNet, which is the best ImageNet performance in 2017 (e.g., DenseNet \cite{DenseNet}). This indicates that recent years of improvement in image classification on ImageNet overfit to ImageNet performance when the downstream task is specified as few-shot classification. We also observe that on datasets like Quick Draw, Traffic Signs, and Ominglot, there is no clear relationship between ImageNet performance and few-shot performance. Since supervised ImageNet performance is usually a strong predictor of other challenging vision tasks,  few-shot classification stands out to be a special task that needs a different and much better generalization ability. Identifying the reasons behind the difference may lead to a deeper understanding of both few-shot classification and vision representation learning.

\keypoint{ImageNet performance is a good predictor of few-shot performance for self-supervised models.} Different from supervised models, for self-supervised models, we observe the clear positive correlation between ImageNet performance and few-shot classification performance. The best self-supervised model only obtains 77\% top-1 accuracy on ImageNet, but obtains more than 83\% average few-shot performance, outperforming all evaluated supervised models. Thus self-supervised algorithms indeed generalize better and the few-shot learning community should pay more attention to the progress of self-supervised learning.




\section{Adaptation Analysis}
\label{sec-test}

In this section, we fix the training algorithm to the CE model trained on \emph{mini}ImageNet and analyze adaptation algorithms.

\subsection{Way and Shot Analysis}
Ways and shots are important variables during the adaptation phase of few-shot classification. For the first time, we analyze how the performance of different adaptation algorithms varies under different choices of ways and shots, with the training algorithm unchanged. For this experiment, we choose ImageNet and Quick Draw as the evaluated datasets because these two datasets have enough classes and images per class to be sampled and are representative for in-domain and out-of-domain datasets, respectively.  For ImageNet, we remove all classes from \emph{mini}ImageNet.


\keypoint{Neural scaling laws for adaptation.} We notice that for Logistic Regression, Finetune, and MetaOPT, the performance curves approximate straight lines when varying the number of shots. This indicates that for the scale of the adaptation dataset, the classification error obeys the traditional neural scaling laws (different from what we found for the scale of the training dataset in Section \ref{sec-train-scale}). While this seems to be a reasonable phenomenon for Finetune, we found it a surprise for Logistic Regression and MetaOpt, which are linear algorithms (for adaptation) built upon frozen features and are thus expected to reach performance saturation quickly.  This reveals that even for small-scaled models trained on \emph{mini}ImageNet, the learned features are still quite linearly-separable for new tasks. However, their growth rates differ, indicating they differ in their capability to scale.

\begin{table*}[t]
\scriptsize
\centering
\setlength\tabcolsep{2pt}
\caption{The average support set size and the degree of task distribution shift of tasks from each dataset on three benchmarks. The metric measuring the degree of task distribution shift is defined by the deviation of feature importance; see Table 3 of \citet{luo2022channel} for details.}
\label{benchmark}
\renewcommand{\arraystretch}{1.2}
    \begin{tabular}{cc|cccc|cccccccccc}
    \hline

    Benchmark & \emph{mini}ImageNet & \multicolumn{4}{c|}{BSCD-FSL} & \multicolumn{10}{c}{Meta-Dataset} \\ 

    Dataset & \emph{mini}ImageNet & ChestX & ISIC & ESAT & CropD & ILSVRC & Omniglot & Aircraft & Birds & Textures & QuickD & Fungi & Flower & Traffic Sign & COCO \\\hline
    Mean support set size & 5 or 25 & \multicolumn{4}{c|}{25 or 100 or 250} & 374.5 & 88.5 & 337.6 & 316.0 & 287.3 & 425.2 & 361.9 & 292.5 & 421.2 & 416.1\\
    
    Task distribution shift & 0.056 & 0.186 & 0.205 & 0.153 & 0.101 & 0.054 & 0.116 & 0.097 & 0.117 & 0.100 & 0.106 & 0.080 & 0.096 & 0.150 & 0.083\\
    \hline
    \end{tabular}
    \vskip -0.05in
\end{table*}

\keypoint{Backbone adaptation is preferred for high-way, high-shot, or cross-domain tasks.} As seen from Figure \ref{test_comparison}, while Finetune and the partial-finetune algorithm TSA do not significantly outperform other algorithms on 1-shot and 5-shot tasks on ImageNet, their advantages become greater when shots or ways increase or the task is conducted on Quick Draw. Thus we can infer that backbone adaptation is preferred when data scale is large enough to avoid overfitting or when the domain shift so large that the learned feature space deforms on the new domain. 

\keypoint{Query-support matching algorithms scale poorly.} Query-support matching algorithms like TSA, MatchingNet, NCC, and URL obtain query predictions by comparing the similarities of query features with support features\footnote{Although these algorithms all belong to metric-based algorithms, there exist other metric-based algorithms like Cosine Classifier that are not query-support matching algorithms. }, different from other algorithms that learn a classifier from the support set directly.  As observed in Figure \ref{test_comparison},  all these algorithms  perform well when the shot is 1 or 5, but scale weaker than power law as the number of shots increases except for TSA on Quick Draw where backbone adaptation is much preferred. Considering that URL as a flexible, optimizable linear head and TSA as a partial fine-tune algorithm have enough capacities for adaptation, their failure to scale well, especially on ImageNet, indicates that the objectives of query-support matching algorithms have fundamental optimization difficulties  during adaptation when data scale increases. 






\subsection{Analysis of Finetune}

As seen from Table \ref{correlation} and Figure \ref{test_comparison}, vanilla Finetune algorithm performs always the best, even when evaluated on in-domain tasks with extremely scarce data. In particular, we have shown that recent partial-finetune algorithms, such as TSA and eTT that are designed to overcome this problem, both underperform vanilla Finetune algorithms. This is quite surprising since the initial Meta-Dataset benchmark \cite{metadataset} shows that vanilla Finetune meets severe overfitting when data is extremely scarce. 

\begin{figure}[t]
\centering
\centerline{\includegraphics[width=1.0\linewidth]{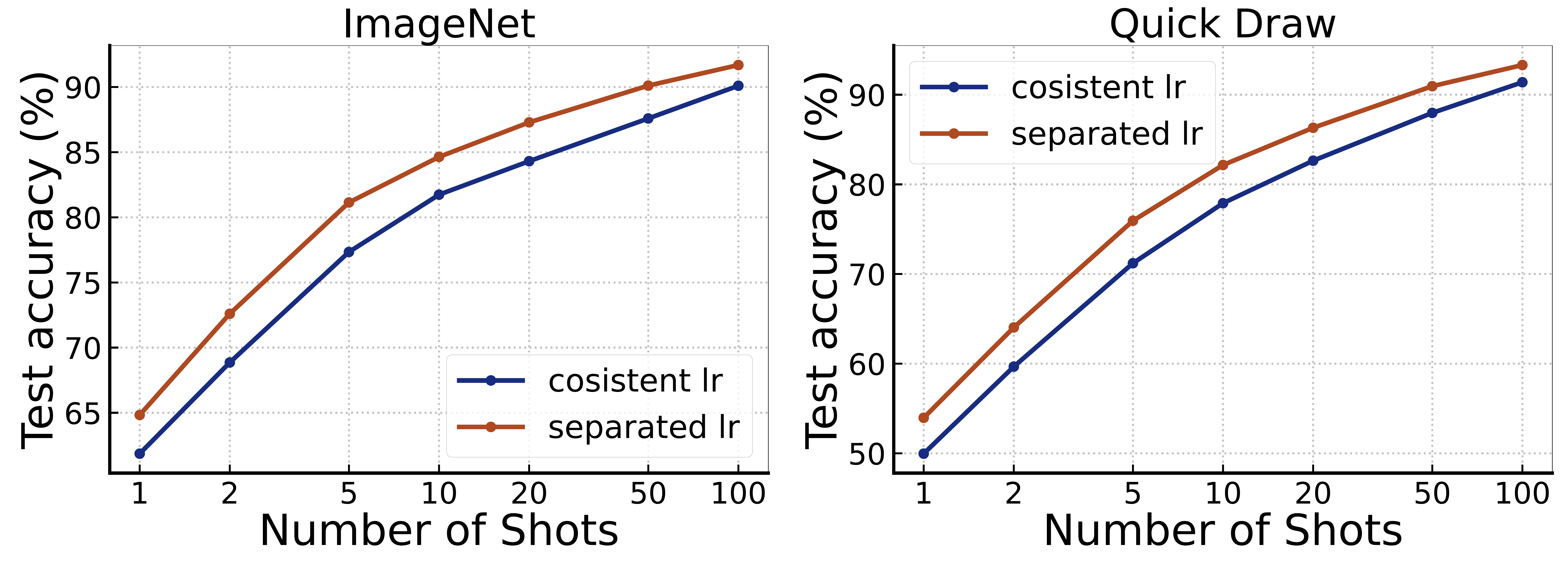}}
\caption{Comparisons of using consistent and separated learning rate for  backbone and linear head during the finetune process.}

\label{single-lr}

\vskip -0.25in
\end{figure}

The reasons lie in two aspects. First, in the original paper of Meta-Dataset, training and adaptation algorithms are bound together, so different adaptation algorithms use different backbones, making it unfair for comparison. This problem is then amplified later in the paper of TSA and eTT, where they use strong backbone for their own adaptation algorithms while copying the original results of Finetune in the benchmark. Second, previous works typically search for a single learning rate for Finetune. We found it important to separately search for the learning rates for backbone and the linear head. This simple change leads to a considerable performance improvement, as shown in Figure \ref{single-lr}. We found that the optimal learning rate of backbone is typically much smaller than that of linear head.

\begin{figure}[t]
\centering
\centerline{\includegraphics[width=0.7\linewidth]{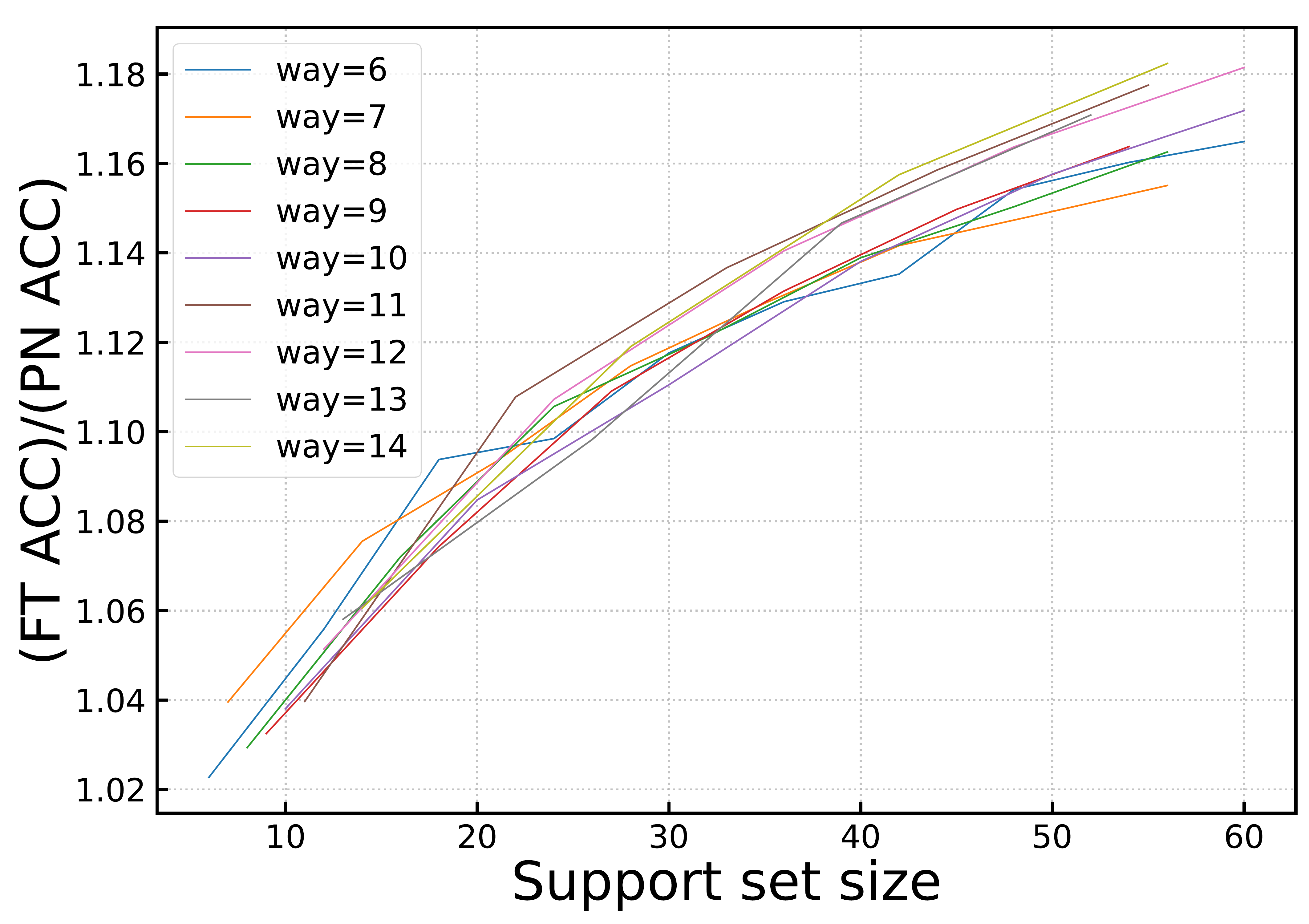}}
\caption{The advantages of Finetune increase as a function of the total number of samples in the support set.}

\label{critical_point}

\vskip -0.3in
\end{figure}

We also wonder what is the critical factor that makes Finetune effective. In Figure \ref{critical_point}, we show how the relative improvement of Finetune over PN changes when we increase total number of samples in the support set (way $\times$ shot). The relative improvement is quite close for all choices of ways, as long as the support set size does not change. Thus support set size is crucial for Finetune to be effective, which aligns with our intuition that the backbone can be adjusted more properly when seeing more data.

\keypoint{Bias of evaluation protocols in different benchmarks.} After analyzing the effectiveness of Finetune, we can now answer a question: why on traditional benchmarks like CIFAR, \emph{mini}ImageNet, \emph{tiered}ImageNet, the state-of-the-art algorithms do not adapt learned backbone during adaptation, but on benchmarks like BSCD-FSL and Meta-Dataset model adaptation becomes popular? As seen from Table \ref{benchmark}, \emph{mini}ImageNet (similarly for CIFAR and \emph{tiered}ImageNet) has a small support set size of 5 or 25 and a small distribution shift from training to test datasets, while BSCD-FSL and Meta-Dataset have 10x larger support set size and encompass datasets with extremely large distribution shift. Thus according to our analysis, backbone adaptation algorithms such as Finetune do not have advantages on benchmarks like \emph{mini}ImageNet, especially when the learning rates are not separated; while on BSCD-FSL and Meta-Dataset, backbone needs adaptation towards new domains and abundant support samples make this possible. To avoid biased assessment, we recommend to the community that, besides reporting standard benchmark results, a method should also report the performance with different, specific ways and shots on datasets with different degrees of distribution shift.

\section{Related Work}
As an active research field, few-shot learning is considered to be a critical step towards building efficient and brain-like machines~\cite{lake2017building}. Meta-learning~\cite{thrun1998learning,schmidhuber1987evolutionary,naik1992meta} was thought to be an ideal framework to approach this goal. Under this framework, methods can be roughly split into three branches: optimization-based methods, black-box methods, and metric-based methods. Optimization-based methods, mainly originated from MAML~\cite{MAML}, learn the experience of how to optimize a neural network given a few training samples. Variants in this direction consider different parts of optimization to meta-learn, including model initialization point~\cite{MAML,LEO,rajeswaran2019meta,zintgraf2019fast,jamal2019task,metaopt}, optimization process~\cite{op-as-model,metafun,metanetwork,metaSGD} or both~\cite{baik2020meta,metacurvature}. Black-box methods~\cite{MANN,CNP,SNAIL,CNAPS} directly model the learning process as a neural network without explicit inductive bias. Metric-based methods~\cite{MatchingNet,PN,relationnet,tapnet,deepemd} meta-learn a feature extractor that can produce a well-shaped feature space equipped with a pre-defined metric. In the context of few-shot image classification, most state-of-the-art meta-learning methods fall into the metric-based and optimization-based ones. 

Recently, a number of non-meta-learning methods that utilize supervised~\cite{closerlook,rethinking,baseline,FLUTE,URL} or unsupervised representation learning methods~\cite{exploring,crosstransformers,Confess,pushing,eTT} to train a feature extractor have emerged to tackle few-shot image classification.  In addition, a bunch of meta-learning methods \cite{metabaseline,deepemd,pushing,ye2020few} learn a model initialized from a pre-trained backbone (Our experiments also consider pretrain+meta-learning training algorithms such as Meta-Baseline, DeepEMD and FEAT. Thus our conclusions hold generally). Since these methods do not strictly follow meta-learning framework, the training algorithm does  not necessarily have a relationship with the adaptation algorithm, and they are found to be simpler and more efficient than meta-learning methods while achieving better performance. Following this line, our paper further reveals that the training and adaptation phases in few-shot image classification are completely disentangled. 

One relevant work~\cite{sbai2020impact} also gives a detailed and comprehensive analysis of few-shot learning, especially on the training process. Our study complements this work in several ways: (1) the neural scaling laws that we found have not been discovered before, which proves the importance of the number of classes in few-shot learning. Although in \citet{sbai2020impact} the significance of the number of classes has also been discussed from different perspectives, there are no clear conclusions in \citet{sbai2020impact} and thus we complement their studies; (2) we observed that larger datasets may lead to degraded performance in specific downstream datasets, both in terms of increasing the number of classes and samples per class. Such findings were not present in \citet{sbai2020impact}, and hence our study opens new avenues for future research by inspecting specific datasets; (3) there's no clear evidence in \citet{sbai2020impact} that simple supervised training scales better than other types of training algorithms; (4) moreover, our paper evaluates 18 datasets, including those beyond ImageNet and CUB, which are the only ones studied in \citet{sbai2020impact}. Thus, our study provides a broader perspective and complements the analysis in \citet{sbai2020impact}.

\section{Discussion}
One lesson learned from our analysis is that training by only scaling models and datasets is not a one-fit-all solution. Either the design of the training objective should consider what the adaptation dataset is (instead of the adaptation algorithm), or the adaptation algorithm should select accurate training knowledge of interest. The former approach limits the trained model to a specific target domain, while the latter approach cannot be realized easily when only few labeled data are provided in the target task which makes knowledge selection difficult or even impossible due to bias of distribution estimation \cite{luo2022channel,xu2022alleviating}. More effort should be put into aligning training knowledge and knowledge needed in adaptation. Although we have shown vanilla Finetune performs so well, we believe that such a brute-force, non-selective model adaptation algorithm is not the final solution, and it has other drawbacks such as having extremely high adaptation cost, as shown in  Appendix \ref{secD}. Viewed from another perspective, our work points to the plausibility of using few-shot classification as a tool to better understand some key aspects of general visual representation learning.

\section*{Acknowledgements}
Special thanks to Qi Yong for providing indispensable spiritual support for the work. We also would like to thank
all reviewers for constructive comments that help us
improve the paper. This work is supported by National Key Research and Development Program of China (No. 2018AAA0102200), and the National Natural Science Foundation of China (Grant No. 62122018, No. U22A2097, No. 62020106008, No. 61872064).

\bibliography{example_paper}
\bibliographystyle{icml2023}

\newpage
\appendix
\onecolumn
\section{Additional Related Work}

\keypoint{Few-shot classification benchmarks.} Earlier benchmarks in few-shot image classification focus on in-domain classification with standard 5-way 1-shot and 5-shot settings, including \emph{mini}ImageNet~\cite{MatchingNet}, FC100~\cite{tadam} and \emph{tiered}ImageNet~\cite{tieredImageNet}. BSCD-FSL benchmark~\cite{guo2020broader} targets at a more realistic cross-domain setting and considers the evaluation of higher shots such as 20 or 50. Meta-Dataset~\cite{metadataset} also targets cross-domain settings, but goes further and considers imbalanced classes and varying numbers of ways and shots. MD+VTAB \cite{md+vtab} further combines Meta-Dataset and VTAB \cite{VTAB} from transfer learning, aiming at connecting few-shot classification with general visual representation learning. 
Although all benchmarks evaluate the models' ability to quickly adapt to new few-shot classification tasks, state-of-the-art methods from different benchmarks differ a lot. In this paper, through a fine-grained test-time analysis, we figure out the reason behind this phenomenon.

\keypoint{Backbone adaptation in few-shot classification.} MAML \cite{MAML} is the first paper that uses Finetune as the adaptation algorithm. However, all hyperparameters of Finetune are fixed before training and the backbone is weak, so MAML does not perform well. Later, Tadam \cite{tadam} designs the first adaptation algorithm that partially adapts the backbone by a black-box meta-learning method. The Baseline algorithm \cite{closerlook} is the first one that uses a combination of non-meta-learning training and Finetune, and achieves surprisingly good results. Another baseline method \cite{baseline} utilizes simple supervised training and Finetune using initialization of the linear layer from feature prototypes. CNAPs \cite{CNAPS} is a partial adaptation meta-learning algorithm that learns on multiple datasets on Meta-Dataset and achieves SOTA results. After CNAPs comes out, several works  emerge that adapt the backbone either by finetuning or partial backbone adaptation in the adaptation phase on Meta-Dataset \cite{FLUTE,TSA,eTT,uppercase,liu2021multi,simpe_CNAPS,FiT}. Our paper reveals that the popularity of this line of research first declined and then increased is related to the bias of evaluation protocols of benchmarks.

\keypoint{Connections of pre-trained models with downstream task performance.} \citet{transfer_better} showed that ImageNet performance has a linear relationship with the downstream transfer performance of classification tasks. Similarly, such a linear relationship was discovered later in domain generalization \cite{natural-distribution-shifts} and open-set recognition \cite{closed-set}. \citet{abnar2021exploring} questioned this result with large-scale experiments and showed that when increasing the
upstream accuracy to a very high number, performance of downstream tasks can saturate. Our experiments on Omniglot and ISIC further ensure this observation, even when the training data is at a small scale. Recently, \citet{entezari2023role} find that the choice of the pre-training data source is essential for the few-shot classification, but its role decreases as more data is made available for fine-tuning, which complements our study.

\section{Details of Experiments}



We reimplement some of the training algorithms in Table \ref{correlation}, including all PN models, all MAML models, CE models with Conv-4 and ResNet-12, MetaOpt, Meta-Baseline, COS, and IER. For all other training algorithms, we use existing checkpoints from  official repositories or the Pytorch library \cite{pytorch}. All reimplemented models are trained for 60 epochs using SGD+Momemtum with cosine learning rate decay without restart. The initial learning rates are all set to 0.1. Training batchsize is 4 for meta-learning models and 256 for non-meta-learning models. The input image size is 84$\times$84 for Conv-4 and ResNet-12 models and 224$\times$224 for other models. We use random crop and horizontal flip as data augmentation at training. Since some models like PN are trained on normalized features, for a fair comparison, we normalize the features of all models for the adaptation phase.

For experiments in Section \ref{sec-train-scale}, to make a fair comparison, we train CE and MoCo for 150 epochs and train PN using a number of iterations that makes the number of seen samples equal. SGD+Momemtum with cosine learning rate decay without restart is used. The backbone used is ResNet-18. Learning rates are all set to 0.1. Training batchsize is 4 for PN and 256 for CE and MoCo.  The input image size is 84$\times$84. During training, we use random crop and horizontal flip as data augmentation for CE and PN, and for MoCo, we use the same set of data augmentations as in the MoCo-v2 paper \cite{moco-v2}. We repeat training 5 times with different sampling of data or classes for each experiment in Section \ref{sec-train-scale}. All pre-trained supervised models in Section \ref{vs-exp} are from Pytorch library, and all self-supervised models are from official repositories.
To avoid memory issues, we only use 500000 image features of the training set of ImageNet for KNN computation of Top-1 ImageNet accuracy for self-supervised models in Section \ref{vs-exp}.

 All training algorithms that are evaluated in this paper, including meta-learning algorithms, set the learnable parameters as the parameters of a feature extractor, and all adaptation algorithms do not have additional parameters that need to be obtained from training. Thus adapting different training algorithms is as easy as adapting different feature extractors with different adaptation algorithms. There exist other meta-learning algorithms \cite{tadam,CNAPS,uppercase,ye2020few,crosstransformers} that meta-learn additional parameters besides a feature extractor, so their training/adaptation algorithms cannot be combined with other adaptation/training algorithms directly. Thus these algorithms are not included in our experiments. One solution is, for each such algorithm, learn the same additional parameters while freezing the backbone for every other trained model, and then 
 we can compare all algorithms. We expect that after doing this the ranking of both training and adaptation algorithms
will still not be changed and we leave it for future work to verify this conjecture.

Throughout the main paper, for all adaptation algorithms that have hyperparameters, we grid search hyperparameters on the validation dataset of \emph{mini}ImageNet and Meta-Dataset. For Traffic Signs which does not have a validation set, we use the hypeparameters averaged over the found optimal hyperparameters of all other datasets. For adaptation analysis experiments in Section \ref{sec-test}, we partition ImageNet and Quick Draw to have a 100-class validation set. The rest is used as the test set.

\section{Additional Tables, Figures, and Analysis}
\subsection{Additional Tables for Section \ref{uncorrelatedness}}
Table \ref{correlation-2} shows 5-way 1-shot results similar to Table \ref{correlation}. Table \ref{correlation-3} and Table \ref{correlation-4} show similar results on \emph{mini}ImageNet. All results lead to the same conclusion that training and adaptation algorithms are uncorrelated. One thing to notice in Table \ref{correlation-2} is that CE model using ViT-base as backbone trained on ImageNet performs particularly well in 1-shot setting. It outperforms DINO in 1-shot setting while underperforms DINO in 5-shot setting. Also, MAML model on Meta-Dataset performs much better than the same model on \emph{mini}ImageNet (possibly due to the use of transductive BN which gives additional unfair flexibility towards new domains). These phenomena show that although training and adaptation algorithms are uncorrelated, the ranking of training algorithms can be influenced by the change of evaluated tasks. Further understanding of how these factors influence the performance of training models is needed in the future.


\begin{table*}[t]
\scriptsize
\centering
\setlength\tabcolsep{5.5pt}
\caption{5-way 1-shot performance of pairwise combinations of a variety of training and adaptation algorithms on Meta-Dataset. We exclude MatchingNet from the adaptation algorithms because MatchingNet equals NCC when the shot is one.}
\label{correlation-2}
\renewcommand{\arraystretch}{1.2}
    \begin{tabular}{cccccccccc}
    \hline
    & & & 
    \multicolumn{7}{c}{Adaptation algorithm} \\ 
    
    Training algorithm & Training dataset & Architecture
                               & MetaOpt                       & NCC                           & LR                            & URL                           & CC                            & TSA/eTT                          & Finetune                 \\
    \hline

    PN  & \emph{mini}ImageNet & Conv-4                        & \acc{38.50}{0.5}          & \acc{38.69}{0.5}          & \acc{38.23}{0.4}          & \acc{38.81}{0.4}          & \acc{38.64}{0.5}          & \acc{41.27}{0.5}          & \acc{42.60}{0.5}            \\
    MAML & \emph{mini}ImageNet   & Conv-4                    & \acc{42.92}{0.5}          & \acc{43.00}{0.5}          & \acc{42.65}{0.5}          & \acc{42.51}{0.5}          & \acc{42.97}{0.5}          & \acc{44.55}{0.5}          & \acc{46.13}{0.5}   \\
    CE   & \emph{mini}ImageNet   & Conv-4                    & \acc{44.49}{0.5}          & \acc{44.88}{0.5}          & \acc{44.88}{0.5}          & \acc{44.48}{0.5}          & \acc{44.82}{0.5}          & \acc{46.20}{0.5}          & \acc{46.92}{0.5}   \\
    MatchingNet   & \emph{mini}ImageNet   & ResNet-12                   & \acc{45.00}{0.5}          & \acc{45.23}{0.5}          & \acc{45.24}{0.5}          & \acc{44.89}{0.5}          & \acc{45.40}{0.5}          & \acc{46.18}{0.5}          & \acc{48.53}{0.5}  \\
    MAML & \emph{mini}ImageNet    & ResNet-12                  & \acc{46.09}{0.5}          & \acc{46.09}{0.5}          & \acc{45.81}{0.5}          & \acc{45.88}{0.5}          & \acc{46.07}{0.5}          & \acc{51.95}{0.5}          & \acc{53.71}{0.5}   \\
    PN       & \emph{mini}ImageNet   & ResNet-12                & \acc{47.32}{0.5}          & \acc{47.53}{0.5}          & \acc{47.33}{0.5}          & \acc{47.53}{0.5}          & \acc{47.65}{0.5}          & \acc{49.36}{0.5}          & \acc{53.06}{0.5}  \\
    MetaOpt  & \emph{mini}ImageNet  & ResNet-12                 & \acc{49.16}{0.5}          & \acc{49.52}{0.5}          & \acc{49.53}{0.5}          & \acc{49.42}{0.5}          & \acc{49.73}{0.5}          & \acc{52.01}{0.5}          & \acc{53.90}{0.5}   \\
    CE      & \emph{mini}ImageNet   & ResNet-12                 & \acc{51.09}{0.5}          & \acc{51.42}{0.5}          & \acc{51.60}{0.5}          & \acc{50.94}{0.5}          & \acc{51.71}{0.5}          & \acc{53.81}{0.5}          & \acc{54.68}{0.5}    \\
    Meta-Baseline     & \emph{mini}ImageNet   & ResNet-12                  & \acc{51.24}{0.5}          & \acc{51.56}{0.5}          & \acc{51.67}{0.5}          & \acc{51.23}{0.5}          & \acc{51.77}{0.5}          & \acc{53.87}{0.5}          & \acc{54.54}{0.5}   \\
    COS      & \emph{mini}ImageNet  & ResNet-12                 & \acc{51.23}{0.5}          & \acc{51.53}{0.5}          & \acc{51.31}{0.5}          & \acc{51.87}{0.5}          & \acc{51.72}{0.5}          & \acc{54.18}{0.5}          & \acc{54.98}{0.5}  \\

        PN     & ImageNet  & ResNet-50                   & \acc{52.50}{0.5}          & \acc{52.84}{0.5}          & \acc{52.71}{0.5}          & \acc{52.90}{0.5}          & \acc{52.93}{0.5}          & \acc{54.34}{0.5}          & \acc{57.40}{0.5}         \\
    IER    & \emph{mini}ImageNet   & ResNet-12                  & \acc{53.31}{0.5}          & \acc{53.63}{0.5}          & \acc{53.82}{0.5}          & \acc{53.24}{0.5}          & \acc{53.98}{0.5}          & \acc{56.32}{0.5}          & \acc{56.98}{0.5}       \\

    Moco v2    & ImageNet   & ResNet-50              & \acc{54.89}{0.5}          & \acc{55.38}{0.5}          & \acc{55.64}{0.5}          & \acc{55.77}{0.5}          & \acc{55.70}{0.5}          & \acc{58.13}{0.5}          & \acc{59.99}{0.5}      \\
    DINO     & ImageNet    & ResNet-50               & \acc{60.81}{0.5}          & \acc{61.37}{0.5}          & \acc{61.61}{0.5}          & \acc{61.96}{0.5}          & \acc{61.81}{0.5}          & \acc{62.69}{0.5}          & \acc{63.61}{0.5}   \\
    CE      & ImageNet   & ResNet-50                 & \acc{62.34}{0.5}          & \acc{62.88}{0.5}          & \acc{62.90}{0.5}          & \acc{63.55}{0.5}          & \acc{63.18}{0.5}          & \acc{65.04}{0.5}          & \acc{65.87}{0.5}   \\
    BiT-S    & ImageNet   & ResNet-50                & \acc{62.41}{0.5}          & \acc{62.95}{0.5}          & \acc{63.15}{0.5}          & \acc{63.40}{0.5}          & \acc{63.40}{0.5}          & \acc{65.02}{0.5}          & \acc{67.05}{0.5}   \\

    CE  & ImageNet & Swin-B & \acc{64.03}{0.5}          & \acc{64.46}{0.5}          & \acc{64.38}{0.5}          & \acc{65.22}{0.5}          & \acc{65.01}{0.5}          & -         & \acc{69.12}{0.5}   \\
    DeiT    & ImageNet & ViT-B   & \acc{64.20}{0.5}          & \acc{64.62}{0.5}          & \acc{64.43}{0.5}          & \acc{65.31}{0.5}          & \acc{65.11}{0.5}          & \acc{66.25}{0.5}                      & \acc{69.12}{0.5}       \\
    DINO    & ImageNet  & ViT-B                  & \acc{64.86}{0.5}          & \acc{65.36}{0.5}          & \acc{65.31}{0.5}          & \acc{66.05}{0.5}          & \acc{65.91}{0.5}          & \acc{67.26}{0.5}                         & \acc{67.89}{0.5}   \\
    CE  & ImageNet  & ViT-B  & \acc{67.19}{0.5}          & \acc{67.61}{0.5}          & \acc{67.56}{0.5}          & \acc{68.00}{0.5}          & \acc{67.85}{0.5}          & \acc{69.78}{0.5}                         & \acc{72.14}{0.4}         \\
    CLIP    & WebImageText  & ViT-B             & \acc{67.95}{0.5}          & \acc{68.68}{0.5}          & \acc{69.10}{0.5}          & \acc{69.85}{0.5}          & \acc{68.85}{0.5}          & \acc{70.42}{0.5}                         & \acc{74.96}{0.5}   \\
    \hline
    \end{tabular}
\end{table*}

\begin{table*}[t]
\scriptsize
\centering
\setlength\tabcolsep{5.5pt}
\caption{5-way 5-shot performance of pairwise combinations of a variety of training and adaptation algorithms conducted on the \emph{mini}ImageNet benchmark.}
\label{correlation-3}
\renewcommand{\arraystretch}{1.2}
    \begin{tabular}{cccccccccc}
    \hline
    & & 
    \multicolumn{8}{c}{Adaptation algorithm} \\ 
    
    Training algorithm & Architecture & MatchingNet
                               & MetaOpt                       & NCC                           & LR                            & URL                           & CC                            & TSA/eTT                         & Finetune                 \\
    \hline
    MAML    & Conv-4                    & \acc{59.80}{0.3} & \acc{57.99}{0.4} & \acc{58.86}{0.2} & \acc{60.93}{0.3} & \acc{60.81}{0.4} & \acc{61.83}{0.3} & \acc{62.40}{0.3} & \acc{62.03}{0.5}   \\
    PN  & Conv-4                        & \acc{63.71}{0.5} & \acc{64.12}{0.5} & \acc{63.67}{0.5} & \acc{65.78}{0.5} & \acc{65.78}{0.4} & \acc{65.82}{0.5} & \acc{65.69}{0.4} & \acc{66.35}{0.5}            \\
    
    CE      & Conv-4                    & \acc{64.09}{0.4} & \acc{66.41}{0.4} & \acc{67.93}{0.3} & \acc{68.92}{0.5} & \acc{68.63}{0.4} & \acc{69.08}{0.5} & \acc{69.22}{0.4} & \acc{69.51}{0.6}   \\
    MatchingNet     & ResNet-12                   & \acc{69.48}{0.3} & \acc{69.71}{0.3} & \acc{69.75}{0.6} & \acc{70.92}{0.4} & \acc{70.86}{0.4} & \acc{71.00}{0.4} & \acc{71.15}{0.2} & \acc{72.31}{0.4}  \\
    MAML    & ResNet-12                  & \acc{70.27}{0.3} & \acc{68.37}{0.6} & \acc{70.09}{0.4} & \acc{71.94}{0.4} & \acc{71.33}{0.3} & \acc{72.10}{0.5} & \acc{75.70}{0.5} & \acc{76.18}{0.3}   \\
    PN       & ResNet-12                & \acc{73.64}{0.4} & \acc{74.03}{0.4} & \acc{74.99}{0.5} & \acc{75.46}{0.4} & \acc{75.72}{0.4} & \acc{75.65}{0.4} & \acc{76.99}{0.3} & \acc{79.62}{0.2}  \\
    MetaOpt   & ResNet-12                 & \acc{75.21}{0.4} & \acc{76.51}{0.5} & \acc{77.69}{0.4} & \acc{78.09}{0.5} & \acc{78.36}{0.4} & \acc{78.43}{0.4} & \acc{80.55}{0.2} & \acc{81.44}{0.2}   \\
    CE      & ResNet-12                 & \acc{76.66}{0.4} & \acc{77.66}{0.4} & \acc{79.97}{0.4} & \acc{80.01}{0.5} & \acc{80.11}{0.5} & \acc{80.34}{0.5} & \acc{80.65}{0.1} & \acc{80.84}{0.2}    \\
    Meta-Baseline       & ResNet-12                  & \acc{77.06}{0.4} & \acc{77.59}{0.4} & \acc{79.85}{0.2} & \acc{80.54}{0.5} & \acc{80.52}{0.4} & \acc{80.77}{0.4} & \acc{80.97}{0.3} & \acc{81.42}{0.2}   \\
    COS    & ResNet-12                 & \acc{79.70}{0.3} & \acc{80.07}{0.4} & \acc{81.01}{0.3} & \acc{81.28}{0.4} & \acc{81.54}{0.4} & \acc{81.52}{0.5} & \acc{81.97}{0.2} & \acc{83.26}{0.2}  \\
    IER   & ResNet-12                  & \acc{80.37}{0.3} & \acc{81.33}{0.3} & \acc{82.80}{0.3} & \acc{83.71}{0.3} & \acc{83.83}{0.3} & \acc{84.04}{0.3} & \acc{83.53}{0.3} & \acc{84.02}{0.2}       \\

    \hline
    \end{tabular}
\end{table*}

\begin{table*}[t]
\scriptsize
\centering
\setlength\tabcolsep{5.5pt}
\caption{5-way 1-shot performance of pairwise combinations of a variety of training and adaptation algorithms conducted on the \emph{mini}ImageNet benchmark.}
\label{correlation-4}
\renewcommand{\arraystretch}{1.2}
    \begin{tabular}{cccccccccc}
    \hline
    & & 
    \multicolumn{7}{c}{Adaptation algorithm} \\ 
    
    Training algorithm & Architecture 
                               & MetaOpt                       & NCC                           & LR                            & URL                           & CC                            & TSA/eTT                         & Finetune                 \\
    \hline
    MAML    & Conv-4                    & \acc{45.97}{0.4} & \acc{46.24}{0.5} & \acc{47.62}{0.5} & \acc{46.81}{0.6} & \acc{47.40}{0.5} & \acc{47.55}{0.4} & \acc{47.40}{0.3}   \\
    PN  & Conv-4                        & \acc{49.79}{0.4} & \acc{50.95}{0.4} & \acc{50.89}{0.4} & \acc{51.01}{0.5} & \acc{50.95}{0.4} & \acc{50.97}{0.3} & \acc{50.65}{0.4}           \\
    
    CE      & Conv-4                    & \acc{51.28}{0.5} & \acc{51.68}{0.7} & \acc{51.07}{0.6} & \acc{52.18}{0.6} & \acc{51.86}{0.7} & \acc{52.88}{0.3} & \acc{51.87}{0.4}   \\
    MatchingNet     & ResNet-12                   & \acc{54.52}{0.5} & \acc{54.96}{0.5} & \acc{54.85}{0.5} & \acc{54.84}{0.6} & \acc{54.89}{0.5} & \acc{55.27}{0.4} & \acc{55.52}{0.4}  \\
    MAML    & ResNet-12                  & \acc{56.43}{0.4} & \acc{55.80}{0.7} & \acc{57.14}{0.7} & \acc{56.06}{0.8} & \acc{57.03}{0.7} & \acc{57.86}{0.4} & \acc{58.49}{0.2}   \\
    PN       & ResNet-12                & \acc{59.91}{0.4} & \acc{60.25}{0.7} & \acc{60.26}{0.7} & \acc{60.01}{0.6} & \acc{60.26}{0.7} & \acc{60.37}{0.5} & \acc{60.67}{0.2}  \\
    MetaOpt   & ResNet-12                 & \acc{60.40}{0.3} & \acc{60.82}{0.5} & \acc{60.40}{0.5} & \acc{61.79}{0.5} & \acc{60.91}{0.5} & \acc{61.89}{0.4} & \acc{62.58}{0.4}   \\
    CE      & ResNet-12                 & \acc{62.53}{0.6} & \acc{62.88}{0.6} & \acc{62.55}{0.6} & \acc{63.15}{0.6} & \acc{62.94}{0.6} & \acc{63.46}{0.4} & \acc{63.33}{0.4}    \\
    Meta-Baseline       & ResNet-12                  & \acc{63.99}{0.2} & \acc{64.92}{0.7} & \acc{64.84}{0.7} & \acc{64.55}{0.7} & \acc{64.91}{0.7} & \acc{64.92}{0.3} & \acc{64.97}{0.2}   \\
    COS    & ResNet-12                 & \acc{64.06}{0.3} & \acc{64.73}{0.9} & \acc{64.71}{0.9} & \acc{64.60}{0.8} & \acc{64.70}{0.9} & \acc{64.92}{0.4} & \acc{65.01}{0.4}  \\
    IER   & ResNet-12                  & \acc{65.05}{0.1} & \acc{66.45}{0.6} & \acc{66.17}{0.6} & \acc{66.68}{0.6} & \acc{66.48}{0.6} & \acc{66.25}{0.3} & \acc{65.86}{0.4}       \\

    \hline
    \end{tabular}
\end{table*}

\subsection{Additional Figures and Analysis for Section \ref{sec-train-scale}}
\begin{figure*}[t]
\centering
\centerline{\includegraphics[width=1.0\linewidth]{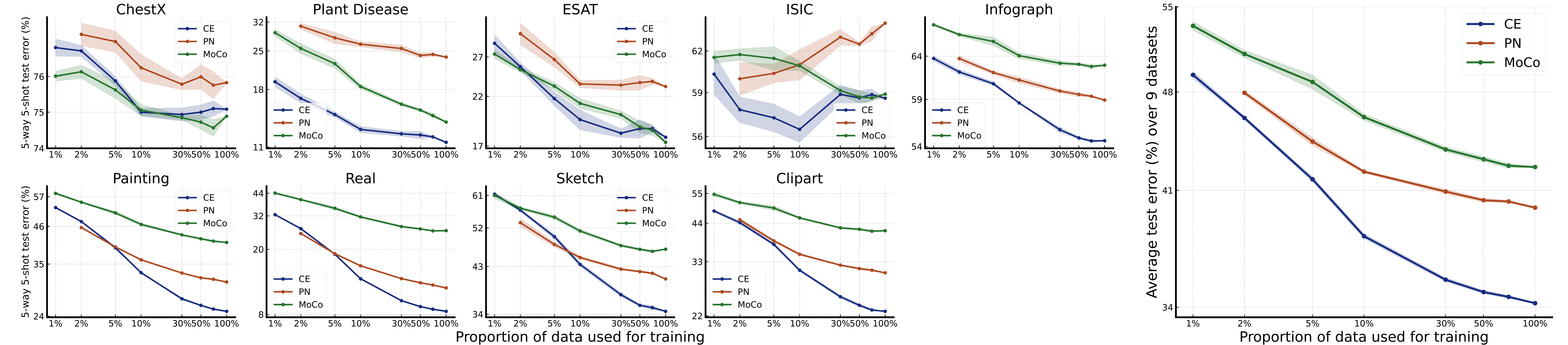}}
\caption{Results of other 9 datasets from BSCD-FSL benchmark and DomainNet about the effect of sample size per training class  on few-shot classification performance. The plot follows Figure \ref{train-dataset-analysis}.}

\label{additional-train-dataset-analysis}
\end{figure*}

\begin{figure*}[b]
\centering
\centerline{\includegraphics[width=1.0\linewidth]{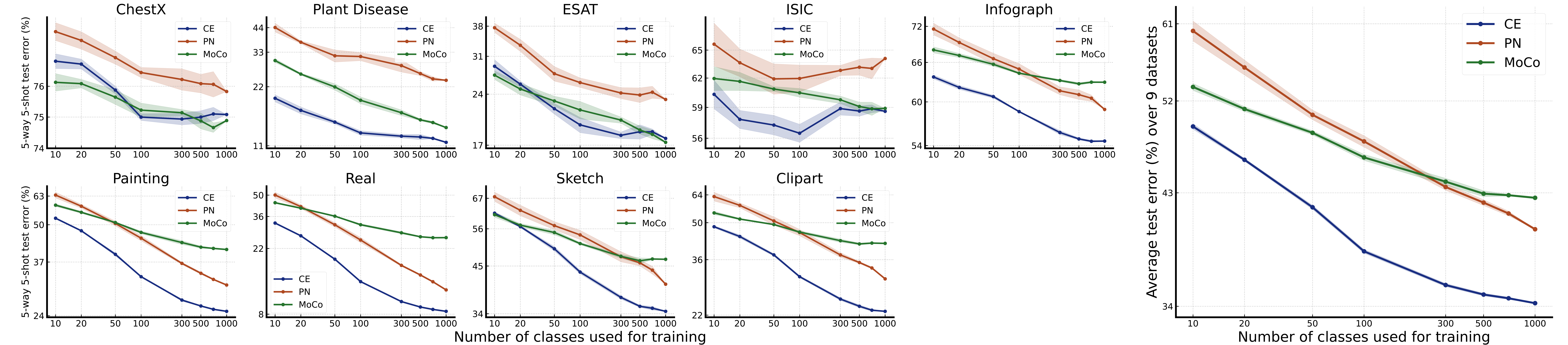}}
\caption{Results of other 9 datasets from BSCD-FSL benchmark and DomainNet about the effect of the number of training classes on few-shot classification performance. The plot follows Figure \ref{train-dataset-analysis-label}.}

\label{additional-train-dataset-analysis-label}

\end{figure*}

Figure \ref{additional-train-dataset-analysis} and Figure \ref{additional-train-dataset-analysis-label} show the data-scaling experiments evaluated on other 9 datasets from BSCD-FSL Benchmark and DomainNet. The general trend is similar to the trend on Meta-Dataset. ISIC shows similar phenomenon to Omniglot in that few-shot performance may not improve if we use larger datasets. But one difference is that for MoCo, few-shot performance does always improve on ISIC, while few-shot performance does not always improve on Omniglot. Also, MoCo performs well on ChestX, while falling behind CE and PN on all other datasets. These show that the knowledge learned from MoCo is somewhat different from that of PN and CE, and this knowledge is useful for classification tasks on ISIC and ChestX. Previous works \cite{Exemplar} has shown that contrastive learning models like MoCo tend to learn more low-level visual features that are easier to transfer. We thus conjecture that low-level knowledge is more important for tasks of some datasets such as ChestX and ISIC. This indicates that the design of the training objective should consider what the test dataset is, so a one-fit-all solution may not exist. We also notice that all datasets in DomainNet except for Quick Draw exhibit similar scale patterns. We know that in DomainNet, each dataset has the same set of classes, while differs in domains. Thus we can infer that the test domain is not the key factor that influences the required training knowledge, but the choice of classes is. In \cite{luo2022channel}, the authors define a new task distribution shift that measures the difference between tasks, taking classes into consideration. It is future work to see whether task distribution shift is the key factor that influences the required training knowledge for each test dataset.

\begin{figure*}[t]
\centering
\centerline{\includegraphics[width=1.0\linewidth]{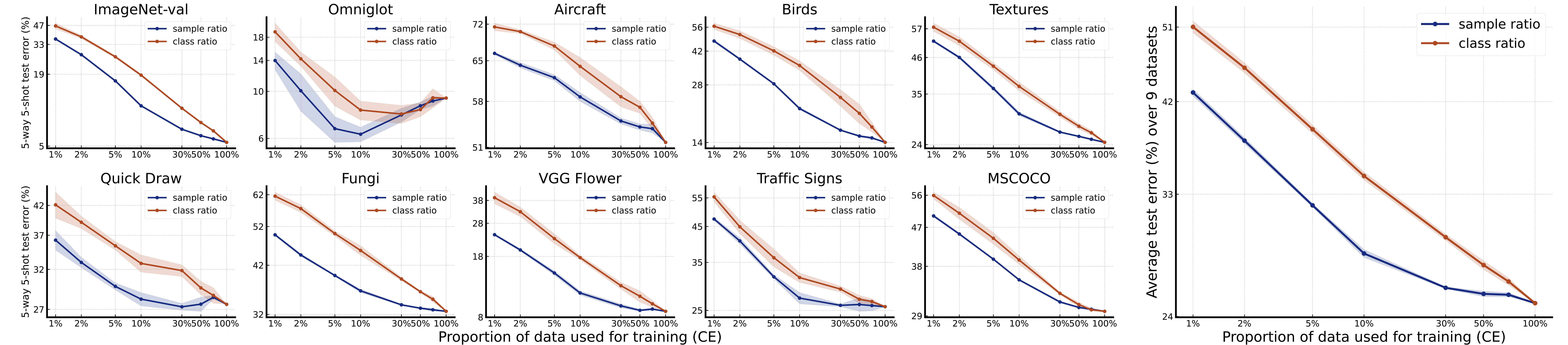}}
\caption{Comparisons of the two data scaling approaches for CE: scaling with sample size per training class and scaling with the number of training classes.}

\label{CE_compare}

\end{figure*}

\begin{figure*}[t]
\centering
\centerline{\includegraphics[width=1.0\linewidth]{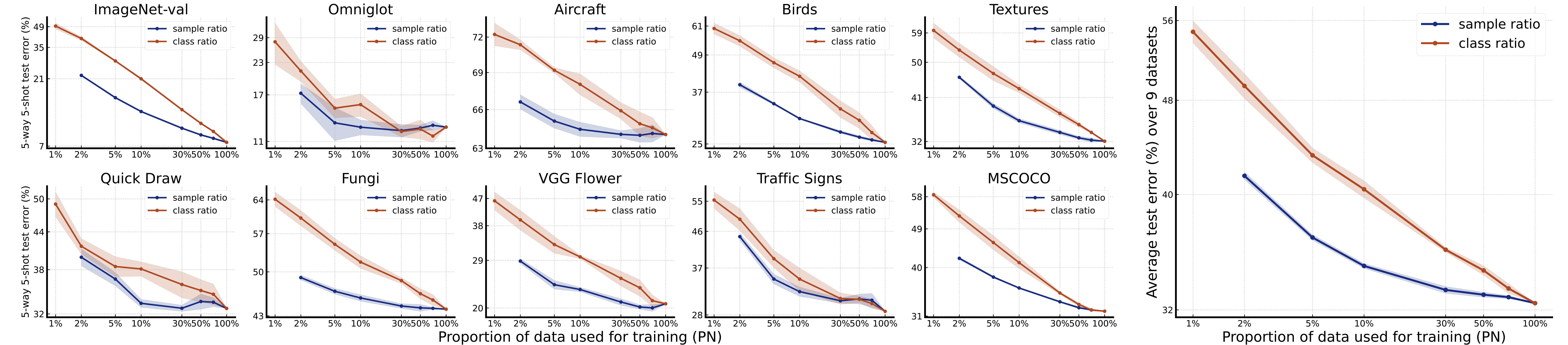}}
\caption{Comparisons of the two data scaling approaches for PN.}
\label{PN_compare}

\end{figure*}

\begin{figure*}[b]
\centering
\centerline{\includegraphics[width=1.0\linewidth]{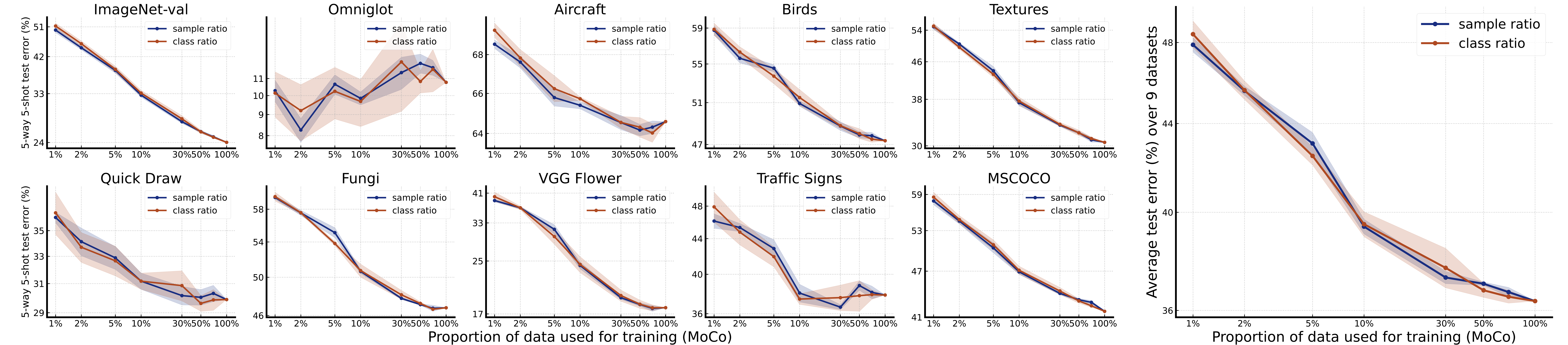}}
\caption{Comparisons of the two data scaling approaches for MoCo.}

\label{MoCo_compare}

\end{figure*}

Figure \ref{CE_compare}-\ref{MoCo_compare} depicts the comparisons of the two data scaling approaches for CE, PN, and MoCo. We can see that for CE and PN, increasing the number of classes is far more effective than increasing the number of samples per class. However, for MoCo, two data scaling approaches present similar performance at every data ratio used for training. Thus we can infer that self-supervised algorithms that do not use labels for supervision indeed do not be influenced by the number of labels. Self-supervised algorithms do not rely on labels, so they treat each sample equally, especially for contrastive learning methods. Thus for self-supervised algorithms, the total number of training samples is the only variable of interest. While this makes self-supervised algorithms particularly suitable for learning on datasets with scarce classes, this also hinders self-supervised algorithms from scaling well to datasets with a large number of classes, e.g., ImageNet-21K or JFM \cite{sun2017revisiting}.

\begin{figure*}[t]
\centering
\centerline{\includegraphics[width=1.0\linewidth]{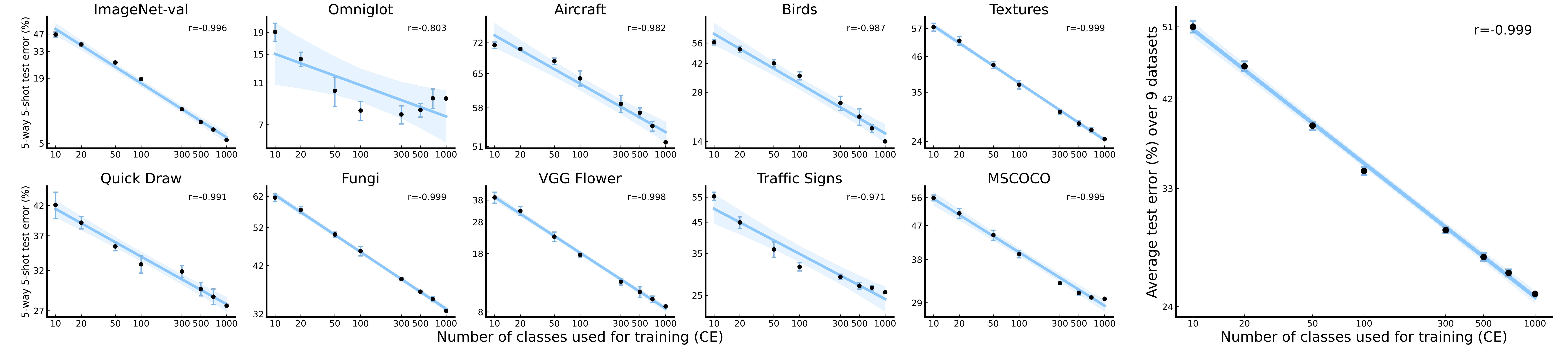}}
\caption{Linear fit of few-shot performance of CE \emph{vs} the number of training classes on logit-transformed axes. ``r'' refers to the correlation coefficient between two axes of data.}

\label{LinearFit_CE}

\end{figure*}

\begin{figure*}[t]
\centering
\centerline{\includegraphics[width=1.0\linewidth]{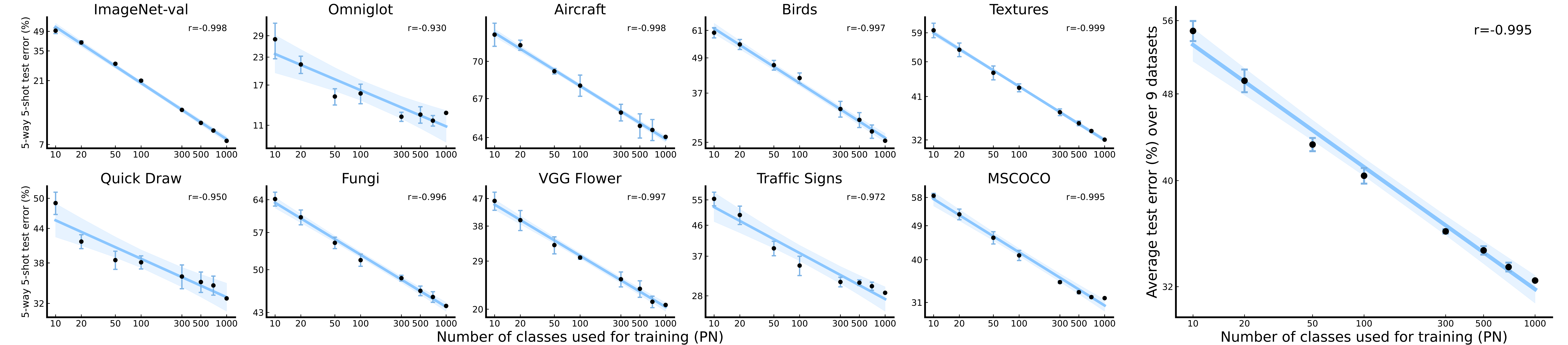}}
\caption{Linear fit of few-shot performance of PN \emph{vs} the number of training classes on logit-transformed axes. ``r'' refers to the correlation coefficient between two axes of data.}

\label{LinearFit_PN}

\end{figure*}

\begin{figure*}[b]
\centering
\centerline{\includegraphics[width=1.0\linewidth]{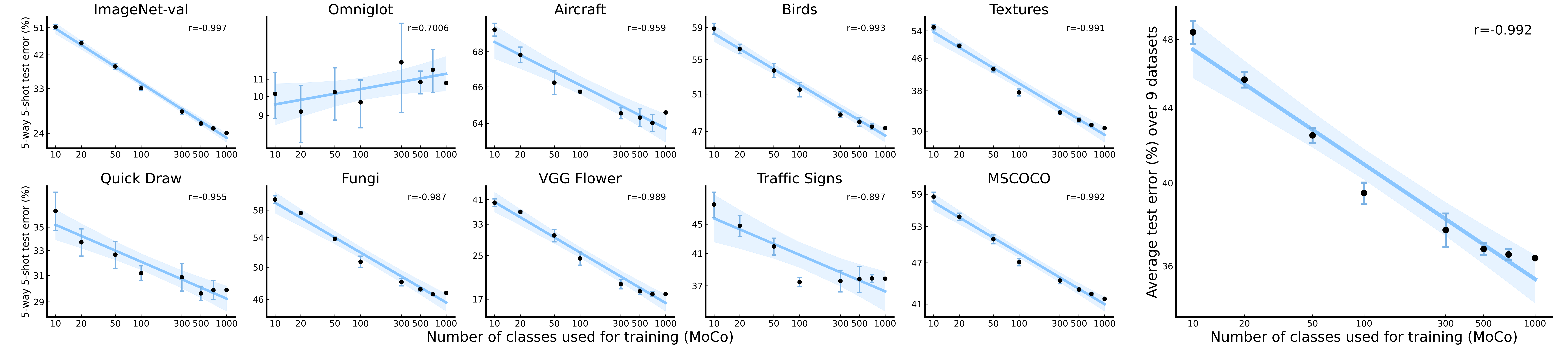}}
\caption{Linear fit of few-shot performance of MoCo \emph{vs} the number of training classes on logit-transformed axes. ``r'' refers to the correlation coefficient between two axes of data.}

\label{LinearFit_MoCo}

\end{figure*}

Figure \ref{LinearFit_CE}-\ref{LinearFit_MoCo} plot the linear fit of few-shot performance vs the number of training classes on logit-transformed axes. We can see that the linear relationship is obvious for most circumstances (most correlation coefficients are larger than 0.9). Thus we have verified the discovery of neural scaling laws wrt the number of training classes. 

\begin{table}[t]
\scriptsize
\centering
\setlength\tabcolsep{2.8pt}
\caption{Detailed results of supervised CE models in Figure \ref{trainnormal_testfsl}. Bold/underline is the best/second best in each column.}
\label{supervised-details}
\renewcommand{\arraystretch}{1.2}
\begin{tabular}{ccccccccccccc}
\hline
Architecture  & ImageNet Top-1 & Avg few-shot & ImageNet-val & Omniglot & Aircraft & Birds & Textures & Quick Draw & Fungi & VGG Flower & Traffic Signs & MSCOCO \\
\hline
ResNet-18     & 68.55          & 79.29        & 96.76        & 92.73    & 59.19    & 90.95 & 79.81    & 70.16      & 73.97 & 94.31      & 78.22         & 74.24  \\
ResNet-34     & 72.50           & 79.18        & 97.66        & 92.76    & 58.65    & 91.71 & 81.57    & 68.57      & 73.54 & 93.80       & 76.27         & 75.77  \\
ResNet-50     & 75.27          & 79.33        & 98.15        & 92.93    & 59.51    & 92.02 & 82.26    & 67.67      & 72.68 & 94.33      & 75.17         & 77.41  \\
ResNet-101    & 76.74          & 79.89        & 98.46        & 92.98    & 60.10     & 92.90  & 81.97    & 69.10       & 74.09 & 94.54      & 75.50          & 77.84  \\
ResNet-152    & 77.73          & 79.02        & 98.62        & 91.33    & 57.20     & 93.36 & 82.36    & 68.12      & 73.85 & 94.26      & 72.37         & 78.37  \\
Swin-T        & 80.74          & 80.86        & 99.14        & 94.17    & 58.26    & 93.40  & \underline{82.70}     & 73.70       & 74.77 & 95.23      & 76.30          & 79.20   \\
Swin-S        & 82.59          & 79.41        & \textbf{99.33}        & 93.17    & 56.94    & 91.89 & 81.07    & 74.14      & 72.01 & 93.25      & 72.68         & \underline{79.58}  \\
Swin-B        & 83.00             & 79.27        & \textbf{99.33}        & 94.87    & 55.26    & 91.25 & 80.63    & 74.54      & 70.71 & 93.99      & 72.32         & \textbf{79.82}  \\
ViT-B         & 80.74          & 80.36        & 98.92        & 94.98    & 58.16    & 92.23 & 80.48    & 73.02      & 71.71 & 93.45      & \textbf{81.33}         & 77.83  \\
ViT-L         & 79.50           & 80.34        & 98.80         & 93.85    & 59.26    & 93.04 & 81.32    & 74.53      & 72.07 & 94.80       & 78.21         & 76.02  \\
DenseNet-121  & 73.60           & 80.78        & 97.52        & 94.88    & 61.62    & 92.89 & 81.62    & 71.95      & 74.30  & 94.73      & 79.58         & 75.49  \\
DenseNet-161  & 76.44          & 81.42        & 98.05        & 93.92    & \textbf{65.87}    & 93.00    & 82.21    & 70.71      & 74.42 & 95.40       & 80.09         & 77.12  \\
DenseNet-169  & 75.07          & 80.65        & 97.78        & 93.60     & 61.71    & 92.43 & 81.77    & 69.55      & 74.28 & 94.98      & 81.21         & 76.29  \\
DenseNet-201  & 75.86          & 81.40         & 97.97        & 94.91    & 61.97    & 93.32 & 82.24    & 73.31      & 73.08 & 95.33      & \textbf{81.33}         & 77.09  \\
RegNetY-1.6GF & 76.01          & \textbf{81.53}        & 97.88        & 94.19    & 62.72    & 93.85 & \textbf{82.84}    & 72.00         & \underline{77.08} & \underline{95.97}      & 77.82         & 77.31  \\
RegNetY-3.2GF & 77.63          & \underline{81.49}        & 98.22        & 93.84    & 63.25    & \textbf{94.07} & \underline{82.70}     & 72.26      & \textbf{77.66} & 95.84      & 75.89         & 77.93  \\
RegNetY-16GF  & 79.39          & 81.21        & 98.57        & 94.82    & 62.16    & \underline{94.02} & 82.46    & 72.34      & 75.79 & 95.68      & 75.03         & 78.62  \\
RegNetY-32GF  & 79.79          & 80.37        & 98.69        & 94.24    & 59.72    & 93.57 & 82.23    & 72.41      & 74.37 & 95.80       & 72.06         & 78.94  \\
RegNetX-400MF & 71.45          & 79.10         & 97.16        & 93.20     & 57.76    & 91.57 & 80.91    & 70.06      & 73.46 & 94.25      & 75.50          & 75.14  \\
RegNetX-800MF & 73.86          & 80.24        & 97.65        & 93.62    & 59.13    & 92.36 & 82.33    & 69.69      & 75.78 & 95.07      & 77.49         & 76.70   \\
MobileNetV2   & 70.54          & 80.90         & 96.86        & 94.26    & 61.03    & 91.87 & 80.61    & 73.30       & 76.13 & 95.56      & 80.64         & 74.70   \\
MobileNetV3-L & 72.91          & 80.48        & 94.71        & 94.91    & 56.63    & 91.45 & 80.68    & \textbf{76.11}      & 74.65 & \textbf{96.49}      & 81.22         & 72.21  \\
MobileNetV3-S & 66.10           & 78.06        & 91.78        & 93.45    & 53.79    & 88.05 & 77.03    & \underline{74.64}      & 72.50  & 94.16      & 80.21         & 68.72  \\
VGG-11        & 67.97          & 75.99        & 93.13        & 93.08    & 54.21    & 85.19 & 78.89    & 65.61      & 70.57 & 93.59      & 72.81         & 69.95  \\
VGG-11-BN     & 69.54          & 77.90         & 93.99        & 94.14    & 58.48    & 87.45 & 81.01    & 64.46      & 73.28 & 94.83      & 76.76         & 70.66  \\
VGG-13        & 68.93          & 76.78        & 93.96        & 93.98    & 54.94    & 87.16 & 79.71    & 66.61      & 70.64 & 93.29      & 73.91         & 70.78  \\
VGG-13-BN     & 70.64          & 78.01        & 94.64        & 92.84    & 58.83    & 88.87 & 81.56    & 64.81      & 74.26 & 94.83      & 74.43         & 71.64  \\
VGG-16        & 70.86          & 77.24        & 95.63        & 92.66    & 55.63    & 89.91 & 79.88    & 62.25      & 72.16 & 93.62      & 76.00            & 73.02  \\
VGG-16-BN     & 72.68          & 78.56        & 96.33        & 91.65    & 60.85    & 91.32 & 81.84    & 62.08      & 74.45 & 93.82      & 76.70          & 74.33  \\
VGG-19        & 71.41          & 77.76        & 96.25        & 94.96    & 57.42    & 90.78 & 79.52    & 64.16      & 71.08 & 91.43      & 76.43         & 74.03  \\
VGG-19-BN     & 73.26          & 79.58        & 96.77        & 92.18    & \underline{64.29}    & 91.80  & 81.57    & 65.23      & 73.43 & 93.15      & 79.82         & 74.80   \\
ConvNeXt-T    & 81.69          & 78.22        & 97.91        & 94.76    & 54.78    & 91.22 & 78.45    & 72.74      & 65.88 & 93.62      & 77.44         & 75.12  \\
ConvNeXt-S    & 82.84          & 77.41        & 98.42        & \underline{95.54}    & 53.58    & 88.60  & 78.18    & 72.53      & 67.26 & 92.63      & 76.10          & 72.29  \\
ConvNeXt-B    & \underline{83.35}          & 77.37        & 98.66        & \textbf{95.65}    & 54.58    & 89.09 & 76.79    & 71.86      & 66.99 & 92.16      & 74.45         & 74.73  \\
ConvNeXt-L    & \textbf{83.69}          & 76.62        & 98.99        & 94.31    & 53.50     & 88.72 & 76.92    & 69.44      & 66.04 & 92.22      & 72.32         & 76.10  \\
\hline
\end{tabular}
\end{table}

\begin{table}[t]
\scriptsize
\centering
\setlength\tabcolsep{1pt}
\caption{Detailed results of self-supervised models in Figure \ref{trainSSL_testfsl}. Bold/underline is the best/second best in each column.}
\label{self-supervised-details}
\renewcommand{\arraystretch}{1.2}
\begin{tabular}{cccccccccccccc}
\hline
Algorithm        & Architecture & ImageNet Top-1 & Avg few-shot & ImageNet-val & Omniglot & Aircraft & Birds & Textures & Quick Draw & Fungi & VGG Flower & Traffic Signs & MSCOCO \\
\hline
BYOL             & ResNet-50    & 62.20           & 77.91        & 92.72        & 92.96    & 52.99    & 80.78 & 83.81    & 73.34      & 70.77 & 96.25      & \underline{81.04}         & 69.30   \\
SwAV             & ResNet-50    & 62.10           & 74.53        & 93.37        & 92.66    & 45.37    & 71.12 & 85.20     & 65.71      & 69.84 & 95.18      & 73.72         & 71.98  \\
SwAV             & ResNet-50-x2 & 62.59          & 74.93        & 92.57        & 94.71    & 45.41    & 68.11 & 85.17    & 68.34      & 70.16 & 95.70       & 75.40          & 71.37  \\
SwAV             & ResNet-50-x4 & 63.65          & 74.60         & 92.40         & 93.89    & 44.99    & 66.26 & 85.71    & 67.71      & 70.16 & 95.53      & 76.38         & 70.80   \\
SwAV             & ResNet-50-x5 & 61.37          & 75.99        & 93.38        & 92.71    & 46.41    & 69.65 & \underline{86.77}    & 67.27      & 72.16 & 96.60       & 79.72         & 72.62  \\
DINO             & ViT-S/8      & \underline{76.94}          & \textbf{83.33}        & 98.16        & 96.61    & \underline{61.20}     & \textbf{95.33} & 85.93    & 73.48      & 80.06 & 98.10       & 80.50          & 78.71  \\
DINO             & ViT-S/16     & 72.48          & 81.52        & 97.28        & 95.01    & 56.88    & \underline{94.80}  & 85.63    & 73.00         & 79.51 & 97.88      & 74.81         & 76.19  \\
DINO             & ViT-B/16     & 74.15          & 81.39        & 97.91        & 95.77    & 50.72    & 92.39 & 86.15    & 73.58      & \textbf{79.77} & \textbf{98.28}      & 78.25         & 77.63  \\
DINO             & ViT-B/8      & 75.74          & \underline{82.85}        & 98.34        & \underline{96.83}   & \textbf{64.67}    & 89.71 & \textbf{87.02}    & 72.39      & 78.83 & \underline{98.21}      & 79.03         & 78.96  \\
DINO             & ResNet-50    & 64.09          & 77.37        & 93.98        & 93.72    & 51.60     & 77.48 & 84.78    & 65.07      & 75.51 & 96.98      & 78.84         & 72.37  \\
MoCo-v1          & ResNet-50    & 41.27          & 67.67        & 87.98        & 88.05    & 41.44    & 61.77 & 77.96    & 61.06      & 61.69 & 89.39      & 62.64         & 65.01  \\
MoCo-v2-200epoch & ResNet-50    & 51.72          & 70.33        & 93.10         & 90.79    & 36.12    & 65.43 & 82.28    & 67.49      & 60.52 & 91.00         & 68.52         & 70.81  \\
MoCo-v2          & ResNet-50    & 59.19          & 71.24        & 94.70         & 89.73    & 34.38    & 70.32 & 84.03    & 66.13      & 61.74 & 91.92      & 70.78         & 72.10   \\
MoCo-v3          & ResNet-50    & 66.61          & 79.95        & 94.91        & 94.61    & 55.45    & 87.31 & 84.75    & 72.27      & 72.75 & 96.68      & \textbf{83.44}         & 72.32  \\
MoCo-v3          & ViT-S        & 65.46          & 76.75        & 94.22        & 93.41    & 45.94    & 84.66 & 83.77    & 73.21      & 69.56 & 94.99      & 72.81         & 72.39  \\
MoCo-v3          & ViT-B        & 69.32          & 78.40         & 95.80         & 94.66    & 47.08    & 85.29 & 84.74    & 75.19      & 72.53 & 96.04      & 76.33         & 73.70   \\
SimSiam          & ResNet-50    & 53.57          & 73.88        & 92.07        & 92.87    & 44.38    & 68.01 & 81.84    & 70.05      & 66.67 & 94.67      & 77.05         & 69.37  \\
Barlow Twins     & ResNet-50    & 63.26          & 77.04        & 93.83        & 92.23    & 49.89    & 79.07 & 84.73    & 68.31      & 71.23 & 96.35      & \underline{81.04}         & 70.53  \\
MAE              & ViT-B        & 20.66          & 46.77        & 39.94        & 93.45    & 26.89    & 35.54 & 33.04    & 72.07      & 30.66 & 52.6       & 41.64         & 35.01  \\
MAE              & ViT-L        & 42.63          & 60.38        & 72.40         & 95.61    & 40.42    & 49.91 & 61.76    & 75.85      & 46.74 & 77.07      & 43.40         & 52.70   \\
MAE              & ViT-H        & 38.50           & 61.43        & 72.32        & 95.36    & 40.96    & 50.97 & 63.64    & 75.11      & 48.91 & 80.02      & 44.64         & 53.27  \\
IBOT             & Swin-T/7     & 73.61          & 81.26        & 97.74        & \textbf{97.05}    & 52.37    & 88.36 & 85.40     & \textbf{77.16}      & 77.05 & 97.46      & 79.16         & 77.37  \\
IBOT             & Swin-T/14    & 74.50           & 81.79        & 97.97        & 96.65    & 51.67    & 93.21 & 85.62    & \underline{76.86}      & \underline{79.64} & 97.75      & 76.90          & 77.83  \\
IBOT             & ViT-S        & 73.12          & 81.25        & 97.54        & 95.67    & 53.97    & 93.91 & 85.32    & 73.77      & 78.23 & 97.66      & 75.82         & 76.86  \\
IBOT             & ViT-B        & 75.28          & 80.16        & 98.04        & 95.8     & 47.01    & 91.57 & 85.21    & 73.78      & 76.57 & 97.81      & 75.63         & 78.02  \\
IBOT             & ViT-L        & 76.37          & 78.59        & 98.27        & 96.18    & 45.60     & 84.78 & 84.02    & 76.27      & 72.93 & 97.18      & 70.92         & \underline{79.46}  \\
EsViT            & ResNet-50    & 69.91          & 75.14        & 97.21        & 88.21    & 42.87    & 80.45 & 84.85    & 62.87      & 70.33 & 95.04      & 75.90          & 75.70   \\
EsViT            & Swin-T       & 74.32          & 81.31        & 97.84        & 96.25    & 50.78    & 94.44 & 85.75    & 74.80       & 78.57 & 97.83      & 75.64         & 77.72  \\
EsViT            & Swin-S       & 76.19          & 79.43        & \underline{98.55}        & 94.93    & 46.50     & 86.50  & 85.52    & 72.77      & 76.41 & 97.15      & 75.71         & 79.33  \\
EsViT            & Swin-B       & \textbf{77.33}          & 77.77     & \textbf{98.77}        & 95.59    & 37.74    & 83.57 & 83.76    & 71.88      & 73.98 & 96.62      & 76.64         & \textbf{80.19}    \\
oBoW             & ResNet-50    & 59.09          & 70.93    & 93.79        & 92.85    & 37.98    & 68.85 & 78.86    & 67.91      & 62.93 & 89.45      & 67.09         & 72.49       \\
InstDisc         & ResNet-50    & 38.13          & 66.85        & 84.70         & 87.18    & 43.25    & 60.72 & 74.23    & 63.84      & 61.34 & 89.54      & 59.42         & 62.14 \\
\hline
\end{tabular}
\end{table}




\subsection{Detailed Results of Figures in Section \ref{vs-exp}}
Table \ref{supervised-details} and Table \ref{self-supervised-details} show the detailed performance of supervised and self-supervised models in Section \ref{vs-exp}.

 \subsection{Additional Analysis for Section \ref{sec-test}}
In Figure \ref{test_comparison}, the few-shot performance is quite close on ImageNet than on Quick Draw. While LR, Finetune, and MetaOPT all follow the power law, their rates are different. All query-support matching algorithms perform similarly to NCC on ImageNet, showing their difficulties to utilize the capacities for generalizing to in-domain tasks. We notice that Cosine Classifier (CC) as a metric-based method performs much better than other metric-based methods when the number of shots is large on ImageNet. This verifies that it is query-support matching that makes algorithms scale poorly, not the use of metric space. We also notice that the behaviors of different algorithms are quite different. While Logistic Regression (LR) performs relatively well when the number of shots increases, its performance quickly drops when the number of ways increases. The ranking of other algorithms such as CC and MetaOPT changes with different situations. It is future work to figure out what influences the performance of these algorithms.

\section{Finetune has High Adaptation Cost}
\label{secD}
For adaptation algorithms like NCC, MatchingNet, Logistic Regression and MetaOPT, all samples of a task just need to go through a single forward pass,  so the adaptation can be very quick, and usually, one task can be completed within one second. For adaptation algorithms like CC and  URL, there is a linear layer that needs to be learned during adaptation, so these methods need several forward and backward pass to update the linear layer. For these algorithms, one task can be completed in several seconds. For adaptation algorithms like Fintune and partial-finetune algorithms such as TSA and eTT, the backward process should be passed through the whole network, and the optimal epoch is usually much higher. So for these algorithms, one task can take from several minutes to several hours to complete, depending on the size of the support set. In practical scenarios, few-shot learning usually requires 
real-time response, so such a long time waiting for one task is 
intolerable.




\end{document}